\documentclass{article} 
\usepackage{iclr2027_conference,times}

\usepackage{amsmath,amsfonts,bm}

\def\eqref#1{equation~\ref{#1}}

\def\1{\bm{1}}

\DeclareMathAlphabet{\mathsfit}{\encodingdefault}{\sfdefault}{m}{sl}
\SetMathAlphabet{\mathsfit}{bold}{\encodingdefault}{\sfdefault}{bx}{n}

\usepackage{url}
\usepackage{amsmath,amssymb}
\usepackage{algorithm}
\usepackage{algpseudocode}
\usepackage{caption}
\usepackage{xcolor}
\usepackage{booktabs}
\usepackage{subcaption}
\usepackage{graphicx}
\usepackage{fontawesome5}
\usepackage{colortbl}
\usepackage{wrapfig}
\usepackage{enumitem}
\usepackage{tabularx} 
\usepackage{multirow}

\definecolor{toolFileEdit}{HTML}{287AE8}
\definecolor{toolLeanRetrieve}{HTML}{F2763B}
\definecolor{toolLeanVerify}{HTML}{174F9C}
\definecolor{toolFileWrite}{HTML}{75ABEE}
\definecolor{toolContentSearch}{HTML}{F5AE80}
\definecolor{toolFileRead}{HTML}{A3AAB5}
\definecolor{toolFileSearch}{HTML}{CF592D}

\newcommand{\tooltag}[2]{%
\textcolor{#1}{\rule{0.7em}{0.7em}}%
\nobreak\hspace{0.3em}\texttt{#2}%
}

\definecolor{mycitecolor}{HTML}{0071bc}
\usepackage{hyperref}
\definecolor{lightblue}{RGB}{230, 240, 245}

\algrenewcommand\algorithmicrequire{\textbf{Require:}}

\title{daVinci-Prover: Agent-Native Mid-Training for Formal Theorem Proving}

\author{%
\textbf{Wenhong Zhu$^{1,2,4}$ \quad Xuefeng Li$^{1,2,4}$ \quad Jiashuo Zhang$^{1,2,4}$ \quad Yu Chen$^{1,2,4}$} \\
\textbf{Tiantian Mi$^{1,3,4}$ \quad Qipeng Guo$^{1,4}$ \quad Pengfei Liu$^{1,2,4}$\thanks{Corresponding author.}} \\[0.5em]
\normalfont $^{1}$\,Shanghai Innovation Institute 
\normalfont $^{2}$\,Shanghai Jiao Tong University 
\normalfont $^{3}$\,Fudan University \\
\normalfont $^{4}$\,Generative Artificial Intelligence Research Lab (GAIR) \\[0.5em]
{\small\hypersetup{hidelinks}%
\href{https://github.com/GAIR-NLP/daVinci-prover}{\faGithub\enspace{github.com/GAIR-NLP/daVinci-prover}}%
}
}

\ifdefined\DaVinciNamedVersion
  \iclrfinalcopy
  \hypersetup{
    pdftitle={daVinci-Prover: Agent-Native Mid-Training for Formal Theorem Proving},
    pdfauthor={Wenhong Zhu, Xuefeng Li, Jiashuo Zhang, Yu Chen, Tiantian Mi, Qipeng Guo, Pengfei Liu}
  }
\fi
\iclrfinalcopy
\ificlrfinal
\fi
\begin{document}

\maketitle
\ificlrfinal
  \fancyhead{}
  \fancyhead[L]{\small\normalfont daVinci-Prover: Agent-Native Mid-Training for Formal Theorem Proving}
  \fancypagestyle{davincifirstpage}{%
    \fancyhead{}%
    \fancyhead[L]{\includegraphics[height=20pt]{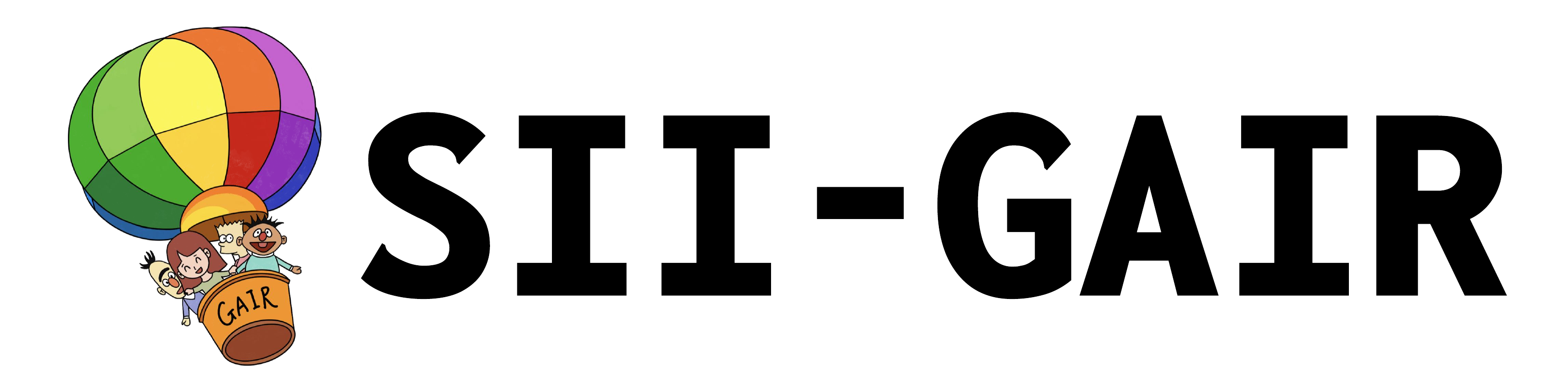}}%
  }
  \thispagestyle{davincifirstpage}
\fi

\begin{abstract}
Advances in large language models (LLMs) are rapidly transforming formal theorem proving. Yet further progress remains constrained by the lack of a transparent and effective data recipe for systematically improving theorem-proving models. In this work, we argue that effective agentic provers require strong \textit{mathematical skill priors} to ground theorem proving and \textit{agentic trajectories} to enable long-horizon, interactive proof construction. To this end, we synthesize an agent-native \textbf{33.7B-token mid-training corpus} covering decomposition, search, self-verification, and revision, together with verifier-grounded agentic trajectories. We further introduce the \textbf{daVinci-Prover} harness, which enables models to construct proofs and interact with a Lean verifier in an isolated workspace, and use it to collect approximately \textbf{36K high-quality agent trajectories} for supervised fine-tuning. Our mid-training recipe yields an average improvement of \textbf{13.7 points} over the matched Qwen3-8B SFT-only baseline across MiniF2F, FATE-M, and ProofNet. When applied to the Qwen3.5 series, our approach also yields strong performance across theorem-proving benchmarks and achieves near-saturating MiniF2F performance with memory compaction. The results highlight agent-native mid-training as an effective foundation for developing stronger interactive theorem-proving agents.
\end{abstract}
\section{Introduction}

Formal theorem proving turns mathematical definitions, statements, and proofs into computational objects that can be checked, reused, and extended. Proof assistants such as Lean~\citep{de2015lean,moura2021lean} provide a particularly attractive interface for language models: they combine the reasoning trajectories required by mathematics with machine-checkable feedback. Recent large language models (LLMs) have made rapid progress in formal proof generation, demonstrating increasingly strong performance across problems ranging from high-school olympiads to research-level mathematics~\citep{chen2025seed,ren2025deepseek,lin2026goedel}.

Existing LLM-based theorem provers broadly follow three paradigms. \textit{(i) Whole-proof generation} produces a complete proof in a single pass, offering a simple and efficient inference procedure~\citep{lin2025goedel}. \textit{(ii) Agentic proving} places the model in an executable environment, where it can inspect the workspace, query libraries, run Lean, interpret verifier feedback, and iteratively revise its proof~\citep{ma2026oprover, wang2026longcat}. \textit{(iii) Orchestrated systems} coordinate multiple reasoning or proving components within structured workflows~\citep{chung2026goedel}. These paradigms are complementary rather than mutually exclusive. Across all three, success depends heavily on the underlying model's ability to decompose the problem effectively, explore alternative proof steps, diagnose failures, and integrate feedback over long reasoning horizons.


Despite this progress, existing formal-reasoning corpora are still dominated by \emph{whole-proof} trajectories, where models primarily observe completed theorem--proof pairs, sometimes augmented with verifier feedback or revised proofs~\citep{lin2025goedel,ma2026oprover}. In contrast, many behaviors essential to effective proving are exposed more explicitly in post-training or inference-time procedures, including decomposing goals into useful sub-theorems~\citep{chen2025seed}, exploring alternative proof steps~\citep{ren2025deepseek, xin2025bfs}, verifying intermediate reasoning~\citep{shao2025deepseekmath}, and localizing and repairing failures. Rather than leaving these capabilities to be acquired primarily during post-training, we argue that they can be injected into mid-training data as reusable \emph{mathematical skill priors}, an idea supported by recent findings on self-improving reasoners~\citep{gandhi2025cognitive}. Beyond these, agentic provers also require a distinct form of training data that captures executable interactions with Lean, including observing proof states, interpreting verifier feedback, and revising failed attempts. These \textit {agentic trajectories} during mid-training can reduce the distribution gap between static model training and agentic inference, where the model operates within an executable proving environment~\citep{zeng2026davinci,yang2026kimi}.

\begin{figure}
    \centering
    \includegraphics[width=1.0\linewidth]{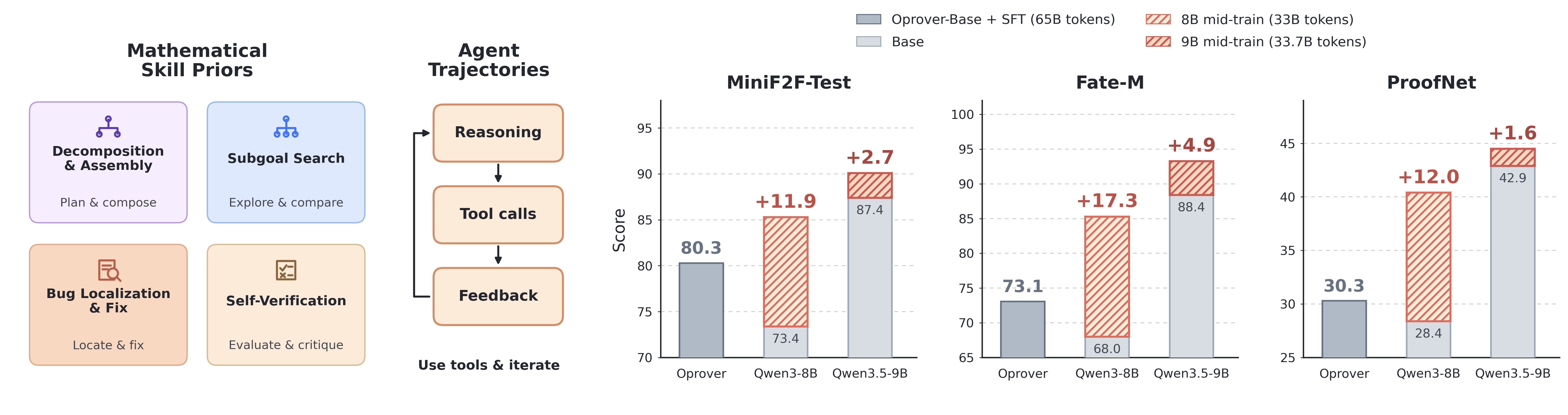}
    \caption{Mid-training data design and benchmark performance. Left: The mid-training data incorporates mathematical skill priors and iterative agent trajectories. Right: Mid-training leads to consistent performance gains on benchmarks across different model families.}
    \label{fig:overview}
\end{figure}



In this work, we introduce \textbf{davinci-Prover}, a family of formal theorem-proving models trained on our curated mid-training corpus and high-quality supervised fine-tuning (SFT) data collected through a lightweight prover harness. As shown in Figure~\ref{fig:overview}, our mid-training corpus covers informal mathematical reasoning, proof decomposition and assembly, verifier-grounded subgoal search, and error diagnosis and revision. We further incorporate agent-native trajectories that expose models to authentic tool calls, execution results, intermediate proof states, and Lean feedback. Together, these data teach the model not only to generate valid proofs, but also to develop the reasoning and interaction patterns required for effective proving in an executable environment. We instantiate this recipe with a 33.7B-token mid-training corpus. For post-training, we build a lightweight prover harness that provides a persistent workspace, typed tools, Lean verification, and trusted completion checks. Using a strong teacher model within this harness, we collect and filter approximately 36K domain-diverse agent trajectories for SFT.
We evaluate the resulting models on various Lean~4 benchmarks. Our main contributions are as follows:

\begin{itemize}[leftmargin=2em]
    \item We construct a 33.7B-token corpus combining mathematical skill priors with environment-native agent trajectories to prepare models for interactive theorem proving.
    \item We develop the daVinci-Prover harness, a lightweight prover harness, and curate approximately 36K high-quality, domain-diverse agent trajectories for supervised fine-tuning. 
    \item Our mid-training improves Qwen3-8B pass@1 by an average of 13.7 points over the matched Base + SFT baseline across MiniF2F, FATE-M, and ProofNet, while consistently improving both Qwen3.5-9B and Qwen3.5-35B-A3B over their corresponding Base + SFT baselines.
\end{itemize}

\section{Background}

Publicly available Lean provers for academic use remain relatively weak~\citep{lin2025goedel, ren2025deepseek}, while stronger systems are either unreleased~\citep{chen2025seed} or prohibitively expensive to run~\citep{varambally2026hilbert, requena2026minimal}. Existing approaches can be broadly categorized into three directions as follows~\citep{chung2026goedel}:

\textbf{(1) Whole-proof generation and refinement.}
These methods generate a complete Lean proof directly from the theorem statement. Representative approaches include DeepSeek-Prover~\citep{ren2025deepseek}, Goedel Prover~\citep{lin2025goedel}, and Kimina-Prover~\citep{wang2025kimina}.
More recent systems augment this paradigm with iterative repair. Goedel-Prover-V2~\citep{lin2026goedel} introduces verifier-guided self-correction, where a complete proof is revised using feedback from the Lean compiler. OProver~\citep{ma2026oprover} further combines whole-proof refinement with retrieval and compiler feedback, and trains the prover directly on such multi-round repair trajectories.

\begin{figure}[htbp]
    \centering
    \includegraphics[width=\linewidth]{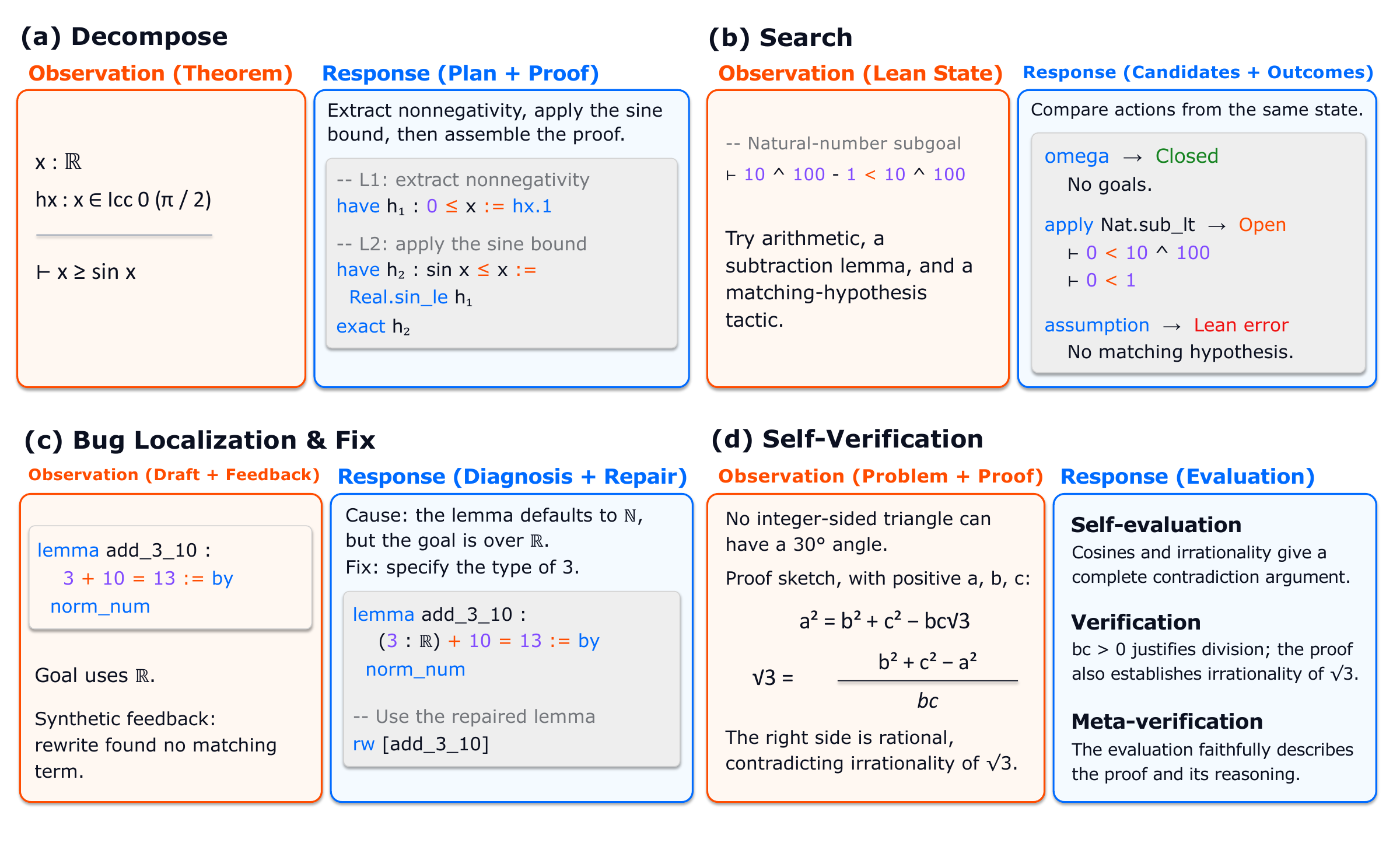}
    \caption{Overview of our midtrain data generation pipeline.}
    \label{fig:data_pipelines}
\end{figure}

\textbf{(2) Agentic provers.} Agentic provers are driven by a single LLM agent that interacts with external tools and adapts its proof strategy based on their outputs. AxProverBase~\citep{requena2026minimal} augments off-the-shelf frontier LLMs with iterative proof refinement, memory, and search capabilities. Numina-Lean Agent~\citep{liu2026numina} similarly builds on Claude Code, exposing Lean verification, theorem retrieval, and auxiliary reasoning tools through MCP, thereby enabling a general-purpose coding agent to carry out formal proof development through iterative tool use.

\textbf{(3) Orchestrated prover systems.}
In contrast to single-agent approaches, orchestrated systems coordinate multiple reasoning or proving components within a predefined workflow, often separating stages such as decomposition, proof sketching, verification, and refinement. Representative systems include Seed-Prover 1.5~\citep{chen2025seed}, Hilbert~\citep{varambally2026hilbert}, and LongCat-Flash-Prover~\citep{wang2026longcat}, which combine multiple stages or specialized components to guide formal proof construction. Goedel-Architect~\citep{chung2026goedel} constructs and iteratively refines a dependency-graph blueprint of formal lemmas, optionally seeded by a natural-language proof, and dispatches the resulting subgoals to tool-equipped Lean provers in parallel.

However, sophisticated orchestrated systems rely on capable agentic provers as their core building blocks; in this work, we focus on advancing the capability of the agentic prover itself. More related work can be found in Appendix~\ref{sec:re}.

\section{Mid-training Data Construction}

\label{mid-train}
This section presents our mid-training data construction framework for developing prover-oriented foundation models. The corpus comprises two complementary components. The first is designed to instill core \textbf{theorem-proving skill priors}, including (1) proof decomposition and assembly, (2) proof search, (3) error diagnosis and revision, and (4) self-verification. The second consists of \textbf{environment-native} agent trajectories that capture authentic interactions with tools, execution outcomes, and verifier feedback.

\subsection{Theorem-proving Skill Priors}

\textbf{Construction of Decomposition and Assembly Trajectories.}
We construct decomposition and assembly trajectories~\citep{ren2025deepseek, zhao2023decomposing, chen2025seed} to develop \textit{lemma-based} proof planning and composition. As shown in Figure~\ref{fig:data_pipelines}(a), a teacher LLM receives a Lean theorem and its verified reference proof, and generates a natural-language proof plan, auxiliary lemmas with dependency information, and a complete assembled Lean proof. The teacher is instructed to specify each lemma's required variables and hypotheses and preserve the original theorem statement. These trajectories are used to identify useful intermediate results, formulate self-contained lemmas, and combine them into a coherent proof.

\textbf{Construction of Verified Subgoal Search Data.}
We replay verified Lean proofs to extract intermediate states and executable snapshots, as shown in Figure~\ref{fig:data_pipelines}(b). For each state, a proposal model generates diverse tactics, which we combine with the reference tactic, anonymize, shuffle, and \textbf{execute independently} from the same snapshot. We retain successful closures, accepted open states, and Lean errors, while excluding timeouts and infrastructure failures. Each trajectory pair links candidate tactics and their search intents with the corresponding execution outcomes, teaching the model to compare alternatives, interpret formal feedback, and distinguish valid intermediate transitions from completed proofs.

\textbf{Construction of Revision Trajectories.} We construct revision trajectories that explicitly connect error diagnosis to proof repair. Given failed proof attempts, diagnostic feedback, and reference corrections, a teacher LLM synthesizes the diagnosis. Figure~\ref{fig:data_pipelines}(c) shows a structured diagnosis with four fields: \textit{location} identifies the affected lemma or proof step; \textit{symptom} describes the observed failure; \textit{cause} explains the underlying issue; and \textit{fix} proposes a targeted correction. Each diagnosis is followed by an explicit revision plan and a revised lemma or proof. These trajectories train the model to localize errors, explain their causes, and formulate targeted corrections.

\textbf{Construction of Informal Reasoning and Verification Trajectories.}
We incorporate informal trajectories from Nemotron-Math-Proofs-v2 using DeepSeekMath-V2-style prompts~\citep{shao2025deepseekmath}. The data covers proof generation with self-evaluation, proof verification, and meta-verification. Proof verification identifies gaps and assesses proof quality, while meta-verification reviews either the generator's self-evaluation or an independent verifier's critique, checking whether the reported defects and assigned scores are justified. These complementary trajectories aim to develop rigorous natural-language reasoning, calibrated self-assessment, and evidence-based critique of both proofs and their evaluations.

\subsection{daVinci-Prover harness}

\begin{wrapfigure}{r}{0.53\linewidth}
    \centering
    \includegraphics[width=\linewidth]
        {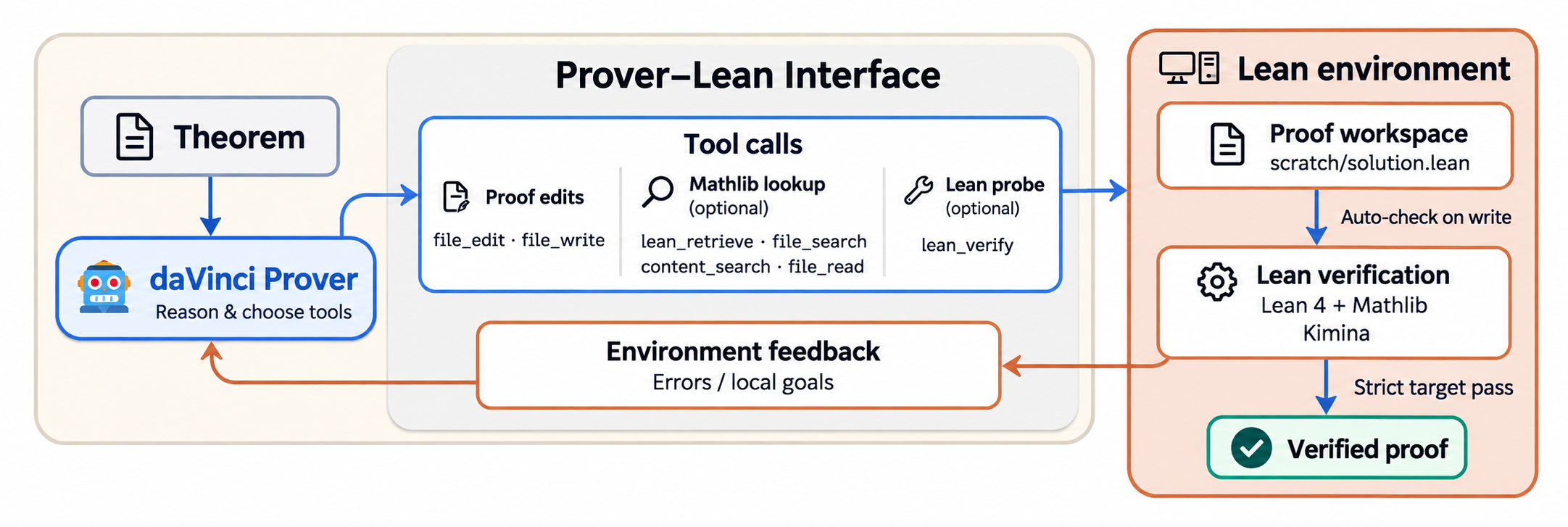}
    \caption{daVinci-Prover agent harness}
    \label{fig:agent-harness}
\end{wrapfigure}

The harness formulates theorem proving as an iterative edit--check--revise loop with Lean4. The input theorem is copied verbatim into \texttt{scratch/solution.lean}, the agent's main proof file. At each turn, the agent selects from seven tools (see Appendix~\ref{tool_schema}) to edit candidates, inspect files, retrieve Lean declarations, and run focused checks. The writable \texttt{scratch/} directory stores proof candidates and auxiliary files, while \texttt{sketch/} holds optional lemma decomposition drafts. The read-only \texttt{target/} directory provides the Lean/Mathlib project used for verification. Every Lean-file edit automatically triggers verification. Draft checks expose local goals and diagnostics, while placeholder-free candidates undergo strict verification. Tool results are added to the conversation to guide subsequent revisions. The episode terminates successfully only when a candidate passes strict Lean verification and checks confirming that its declaration and elaborated type match the original theorem.

\subsection{The Final Dataset: Davinci-prover 33.7B Collection}

\textbf{Data Source.}
We use a curated corpus drawn from \texttt{NuminaMath-LEAN}~\citep{wang2025kimina}, AoPS~\citep{aops2026}, Lean Workbook~\citep{ying2024lean}, \texttt{Goedel LM/SFT\_dataset\_v2}~\citep{lin2025goedel}, and OProofs~\citep{ma2026oprover}. To mitigate data contamination~\citep{zhu2024clean}, we remove duplicate problems and exclude examples that overlap with our evaluation benchmarks. These filtering steps remove 0.03427\% of the 33.7B-token mid-training corpus, with token counts computed using the Qwen3.5 tokenizer; see Appendix~\ref{decontamination}.

\textbf{Data Cleaning and Reconstruction.}
Following Section~\ref{mid-train}, we use DeepSeek-V4-Flash and DeepSeek-V4-Flash-0731~\citep{xu2026deepseek} to construct an approximately 29B-token mid-training corpus, whose composition is shown in Table~\ref{tab:mid-training-data}. We remove repetitive system prompts, tool schemas, and source-specific boilerplate while retaining the problems, reasoning traces, tool interactions, and outcomes. Each example is then reconstructed into a unified training document and semantically chunked to at most 32K tokens at natural boundaries.

\begin{wraptable}{r}{0.6\textwidth}
\centering
\vspace{-8pt}
\caption{Categories and statistics of the mid-training corpus.}
\label{tab:mid-training-data}
\setlength{\tabcolsep}{4pt}
\renewcommand{\arraystretch}{1.05}
\resizebox{\linewidth}{!}{%
\begin{tabular}{lccc}
    \toprule
    \textbf{Category} & \textbf{\# Samples (M)} & \textbf{\# Toks. (B)} & \textbf{Share (\%)} \\
    \midrule
    \rowcolor{orange!10}
    \textbf{Pre-Replay Data} & \textbf{8.131} & \textbf{28.6} & \textbf{84.8} \\
    \quad Self-verfication & 0.083 & 5.70 & 16.9 \\
    \quad Theorem Decomposition & 4.662 & 8.43 & 25.0 \\
    \quad Proof Revision & 1.027 & 3.84 & 11.4 \\
    \quad Proof Search & 1.962 & 3.45 & 10.2 \\
    \quad Environment-Aware Agent Trajectories & 0.397 & 7.17 & 21.3 \\
    \midrule
    \rowcolor{orange!10}
    \textbf{Replay Data}   & \textbf{1.454}   & \textbf{5.11}  & \textbf{15.2} \\
    \midrule
    \rowcolor{orange!10}
    \textbf{Total}  & \textbf{9.585}  & \textbf{33.70}  & \textbf{100.0} \\
    \bottomrule
\end{tabular}%
}

\vspace{2pt}
\vspace{-8pt}
\end{wraptable}

\textbf{Data Replay.}
We define a parent example as one complete logical trajectory before 32K semantic chunking. We assign each source a replay budget according to the desired data mixture and select high-quality parent examples using source-specific scores. To preserve diversity, we cover distinct problems before selecting additional trajectories from the same problem. Once a parent example is selected, all of its semantic chunks are replayed together and processed using the same reconstruction and packing pipeline as the original corpus. This adds 5.11B replay tokens, resulting in a final corpus of 33.7B tokens.

\section{Supervised Fine-Tuning}


\begin{wrapfigure}{r}{0.53\linewidth}
    \centering
    \includegraphics[width=\linewidth]
        {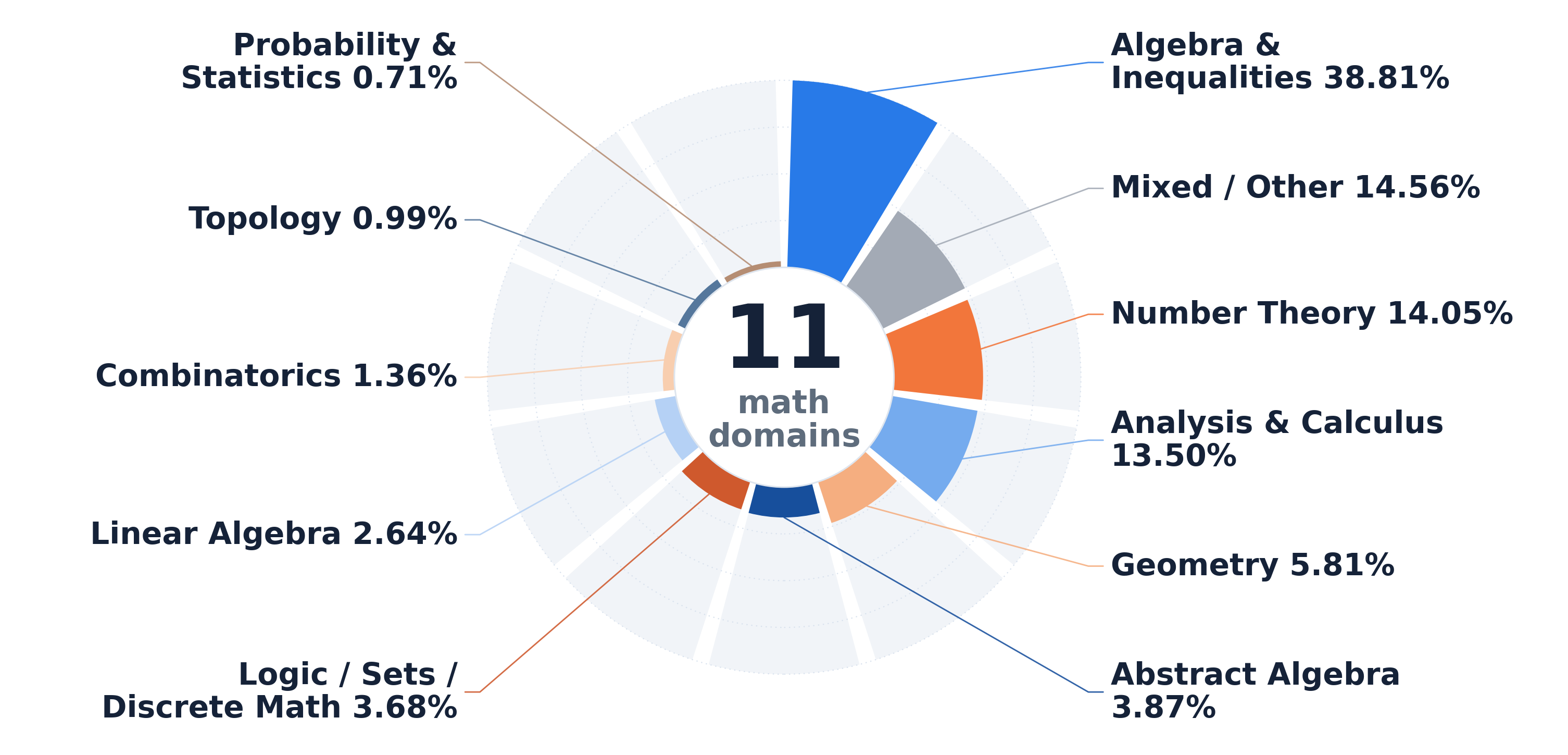}
    \caption{SFT Data Overview}
    \label{fig: sft_data}
\end{wrapfigure}

\textbf{Data Source.}
Our SFT training corpus consists of trajectories generated through harness rollouts. We sample problems from the \texttt{Goedel-LM/RL\_dataset\_V2} dataset~\citep{lin2025goedel}, which contains approximately 99K unverified problems, and use \texttt{Goedel-LM/SFT\_dataset\_V2} for domain supplementation.  We limit our interaction turns to 30 turns with a maximum of 200 tool calls. The trajectories are collected using DeepSeek-V4-Flash-0731.


\textbf{SFT Data Overview.}
We ultimately collect approximately 36K high-quality SFT trajectories from successful theorem-proving runs. The resulting dataset spans a broad range of mathematical domains, including algebra, number theory, geometry, analysis, combinatorics, topology, probability and statistics, linear algebra, abstract algebra, and logic and set theory, as shown in Figure~\ref{fig: sft_data}. Following the filtering strategy of~\citet{raoof2026openthoughts}, we remove low-quality trajectories exhibiting undesirable behaviors, such as excessive repetition or unnecessary tool calls, as well as trajectories containing fewer than 5 turns. We further retain only trajectories that fit within a 144K-token context window.



\section{Experiments}
\textbf{Models.} We select  \texttt{Qwen3.5-9B-Base} and
\texttt{Qwen3.5-35B-A3B-Base} for experiments. We also perform ablation on \texttt{Qwen3-8B-Base}, see Section~\ref{corpora_comparison}.

\begin{table*}[hpt]
\centering
\small
\setlength{\tabcolsep}{5pt}
\renewcommand{\arraystretch}{1.15}
\caption{Theorem-proving performance. Best results are in \textbf{bold}, and second-best results are \underline{underlined}. $\dagger$ marks scores not re-evaluated by us under our protocol.}
\resizebox{\textwidth}{!}{
\begin{tabular}{l c c c c c c}
\toprule
\textbf{Model} & \textbf{\#Params} & 
\textbf{MathOlympiad} & 
\textbf{MiniF2F} & 
\textbf{ProofNet} & 
\textbf{ProverBench} & 
\textbf{Putnam} \\ 
\midrule

\multicolumn{7}{c}{\textit{Open-Weights Reasoning Models (\textbf{pass@32})}} \\
\midrule
DeepSeek-V3.2              & 671B-A37B & 14.7$^\dagger$ & 77.9$^\dagger$ & 20.4$^\dagger$ & 42.8$^\dagger$ & 5.8$^\dagger$ \\
Kimi-K2.5                  & 1T-A32B   & 7.5$^\dagger$  & 76.6$^\dagger$ & 19.9$^\dagger$ & 44.3$^\dagger$ & 1.2$^\dagger$ \\
\midrule

\multicolumn{7}{c}{\textit{Open-Weights Prover Models (\textbf{pass@32})}} \\
\midrule
DeepSeek-Prover-V2-7B      & 7B   & 7.9$^\dagger$  & 73.9$^\dagger$ & 18.1$^\dagger$ & 45.7$^\dagger$ & 1.5$^\dagger$ \\
DeepSeek-Prover-V2   & 671B-A37B  & 13.9$^\dagger$ & 82.4$^\dagger$ & 30.5$^\dagger$ & 52.9$^\dagger$ & 3.3$^\dagger$ \\
Goedel-Prover-V2-8B        & 8B   & 9.9$^\dagger$  & 79.4$^\dagger$ & 17.3$^\dagger$ & 48.6$^\dagger$ & 0.7$^\dagger$ \\

Goedel-Prover-V2-32B       & 32B  & 16.0$^\dagger$ & 85.8$^\dagger$ & 22.0$^\dagger$ & 51.0$^\dagger$ & 5.0$^\dagger$ \\

LongCat-Flash-Prover       & 560B-A27B & 16.9$^\dagger$ & 84.4$^\dagger$ & 19.9$^\dagger$ & 49.9$^\dagger$ & 4.9$^\dagger$ \\
\quad whole-proof w/ TIR   & 560B-A27B &  {27.5}$^\dagger$ & {90.2}$^\dagger$ &  {36.1}$^\dagger$ & {57.9}$^\dagger$ & {10.4}$^\dagger$ \\
\quad sketch-proof w/ TIR   & 560B-A27B & {35.8}$^\dagger$ & {93.9}$^\dagger$ & {47.3}$^\dagger$  & {66.5}$^\dagger$ &  \underline{28.9}$^\dagger$ \\

OProver-8B                 & 8B   & 21.7$^\dagger$ & {91.8}$^\dagger$ & 31.9$^\dagger$ & 56.0$^\dagger$ & 9.0$^\dagger$ \\
OProver-32B                & 32B  & 22.8$^\dagger$ & 93.3$^\dagger$ & 33.2$^\dagger$ & 58.2$^\dagger$ & 11.3$^\dagger$ \\
\midrule

\multicolumn{7}{c}{\textit{daVinci-Prover Harness (\textbf{pass@8})}} \\
\midrule
DeepSeek-V4-Flash-0731 & 284B-A13B   & \textbf{52.2} & \textbf{95.9} & \textbf{56.5} & \textbf{68.9}  & \textbf{39.7} \\
Qwen3.5-9B-Base + SFT & 9B &  32.3 &   92.6 & 51.1 & 66.2  & 22.3\\
Qwen3.5-9B-Midtrain + SFT & 9B & 35.3  &  93.9 &  52.7 &  66.5 &  24.9 \\
Qwen3.5-35B-Base + SFT & 35B-A3B & 35.8 & 93.0  &  52.7  & 66.5 & 24.3   \\
Qwen3.5-35B-Midtrain + SFT & 35B-A3B & \underline{36.1} &  \underline{94.3}  & \underline{53.8} &  \underline{67.1} &  26.5\\
\bottomrule
\end{tabular}
}
\label{tab:main_results}
\end{table*}

\textbf{Benchmarks.}
We evaluate on five Lean~4 theorem-proving benchmarks spanning diverse levels of mathematical difficulty and scope: MiniF2F~\citep{zheng2021minif2f}, which contains 244 high-school olympiad problems; MathOlympiadBench~\citep{lin2025goedel}, consisting of 360 recent competition problems; ProofNet~\citep{azerbayev2023proofnet}, featuring 186 undergraduate-level theorems; ProverBench~\citep{ren2025deepseek}, which includes 325 problems covering both olympiad and undergraduate mathematics; and PutnamBench~\citep{tsoukalas2024putnambench}, a collection of 672 Putnam competition problems representing the most challenging benchmark in the suite. We additionally evaluate on FATE-M and FATE-H~\citep{shen2025real}, which consist of standard textbook exercises and challenging problems from abstract and commutative algebra.

\textbf{Evaluation.}
We evaluate all models using the daVinci-Prover harness with a temperature of 1.0, the context length of 200K, and up to 30 interactions. We report pass@1 averaged over three runs where applicable, and pass@8 otherwise. All experiments use Lean v4.26 and Mathlib at commit \texttt{2df2f0150c275ad53cb3c90f7c98ec15a56a1a67}. We reject any proof attempt that introduces axioms or local instances, as these could circumvent formal verification, including through \texttt{native\_decide}.

\subsection{Effectiveness of daVinci-Prover Harness}

\begin{wraptable}{r}{0.58\textwidth}
  \centering
    \caption{Effectiveness of the daVinci-Prover harness. We report pass@1 averaged over three runs. ``w/ ret.'' denotes the use of the mathlib retrieval tool.}
  \label{tab:harness_performance}
  \resizebox{\linewidth}{!}{
  \begin{tabular}{lcccc}
    \toprule
    Model & MiniF2F & FATE-M & FATE-H & Putnam \\
    \midrule
    Qwen3.6-35B-A3B
      & 68.0 & 48.7 & 4.0 & 3.0 \\
    Qwen3.6-35B-A3B (w/ ret.)
      & 73.0 & 57.3 & 5.0 & 3.4 \\
    DeepSeek-V4-Flash
      & 90.2 & 94.7 & 19.0 & 18.9 \\
    DeepSeek-V4-Flash (w/ ret.)
      & \textbf{91.3} & \textbf{99.3} & \textbf{29.0} & \textbf{22.8} \\
    \bottomrule
  \end{tabular}
  }
\end{wraptable}

We select DeepSeek-V4-Flash-0731 and Qwen3.6-35B-A3B to evaluate the harness's effectiveness.  
As shown in Table~\ref{tab:harness_performance}, we evaluate our lightweight agent harness across different model backbones, with high reasoning effort enabled for DeepSeek-V4-Flash-0731. Despite its simple design, augmenting the harness with a mathlib retrieval tool consistently improves performance across benchmarks of varying difficulty. The gains are particularly pronounced on the \textbf{more challenging} Fate-H and Putnam benchmarks. We observe a similar improvement trend with Qwen3.6-35B-A3B, indicating that the benefit of retrieval is not specific to a single model backbone.

\subsection{Agentic Prover Performance}
The overall performance results are presented in Table~\ref{tab:main_results}. We provide the following analysis:

\textbf{Agentic prover outperforms existing whole-proof baselines.}
 DeepSeek-V4-Flash-0731, evaluated with the daVinci-Prover harness, achieves the highest reported scores across all five benchmarks. Its interactive pass@8 results exceed the reported pass@32 results of whole-proof generation and refinement methods. The results highlight the potential of agentic proving to enable strong theorem proving with whole proof.

\textbf{High-quality SFT trajectories effectively elicit theorem-proving capabilities.}
We further use the proof trajectories generated by the daVinci-Prover harness as SFT data to train Qwen3.5 models. Despite their substantially smaller model sizes, the resulting models achieve strong performance across all benchmarks, indicating that trajectories collected by the harness provide effective supervision for transferring its theorem-proving behavior to smaller models.

\textbf{Mid-training consistently improves performance over the base models.}
Across both model scales, the Midtrain + SFT variants consistently outperform their corresponding Base + SFT counterparts. The gains are particularly pronounced on more challenging benchmarks such as MathOlympiad and Putnam. For Qwen3.5-9B, mid-training improves MathOlympiad from 32.3 to 35.3 and Putnam from 22.3 to 24.9, while maintaining consistent improvements on the other benchmarks. A similar trend is observed for Qwen3.5-35B. These results suggest that mid-training provides a stronger initialization for subsequent theorem-proving SFT, with particularly clear benefits on more difficult mathematical reasoning tasks.

\section{IN-DEPTH ANALYSES}

\subsection{Training Process}

\begin{figure}[htbp]
    \centering

    \begin{subfigure}{0.32\linewidth}
        \centering
        \includegraphics[width=\linewidth]{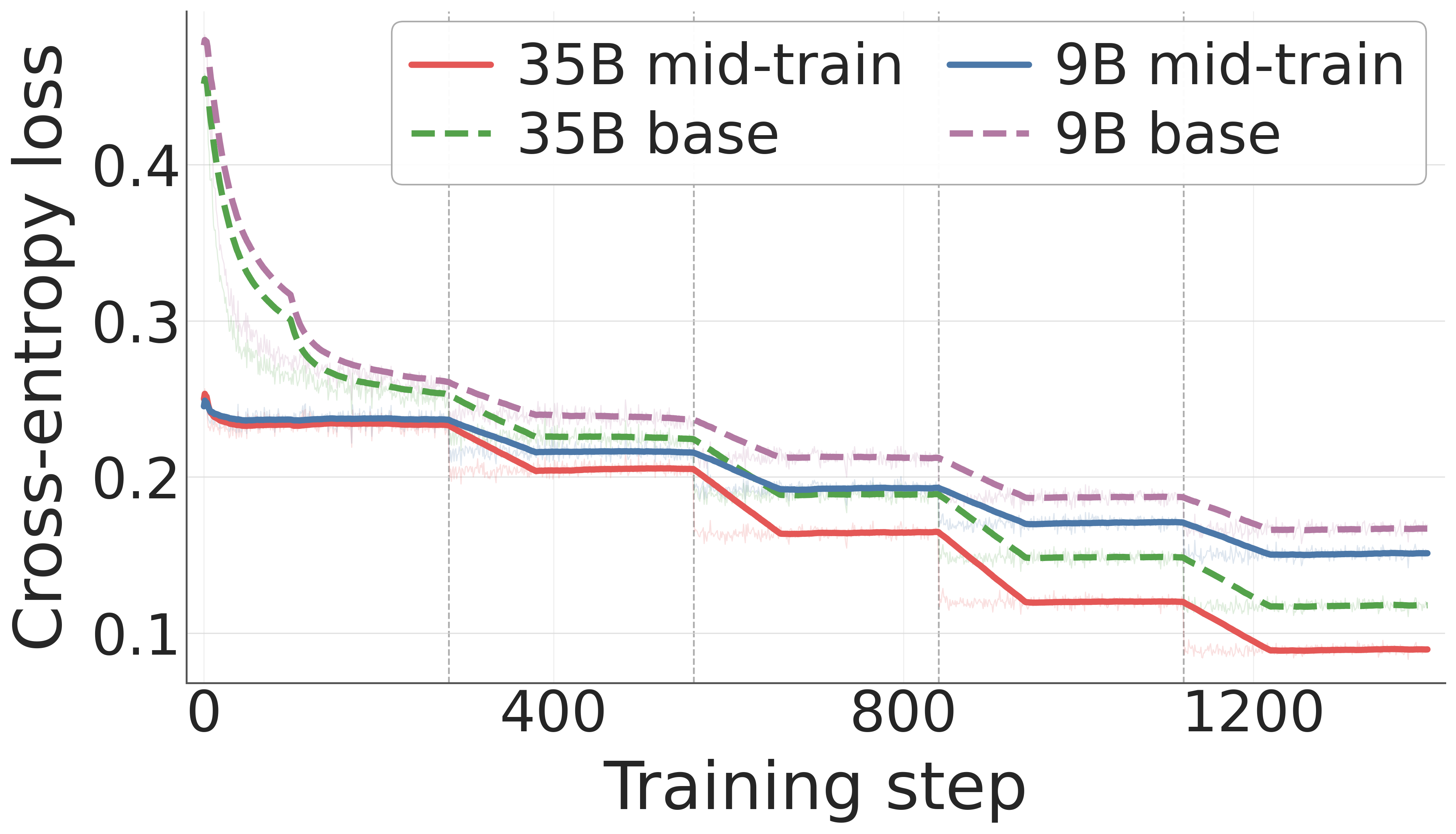}
        \caption{SFT training loss}
        \label{fig:sft_loss}
    \end{subfigure}
    \hfill
    \begin{subfigure}{0.32\linewidth}
        \centering
        \includegraphics[width=\linewidth]{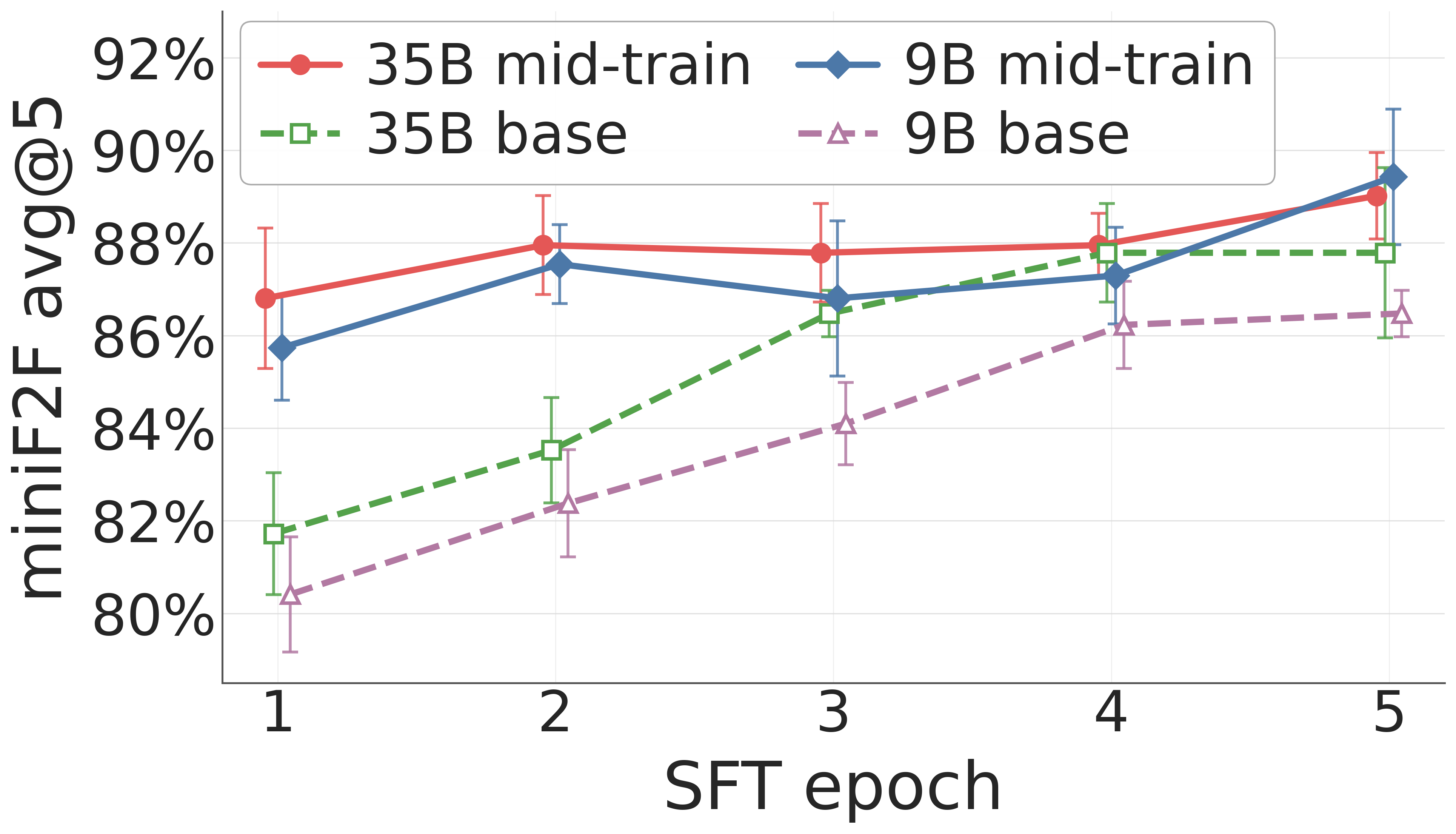}
        \caption{miniF2F $\mathrm{avg}@5$}
        \label{fig:sft_performance}
    \end{subfigure}
    \hfill
    \begin{subfigure}{0.32\linewidth}
        \centering
        \includegraphics[width=\linewidth]{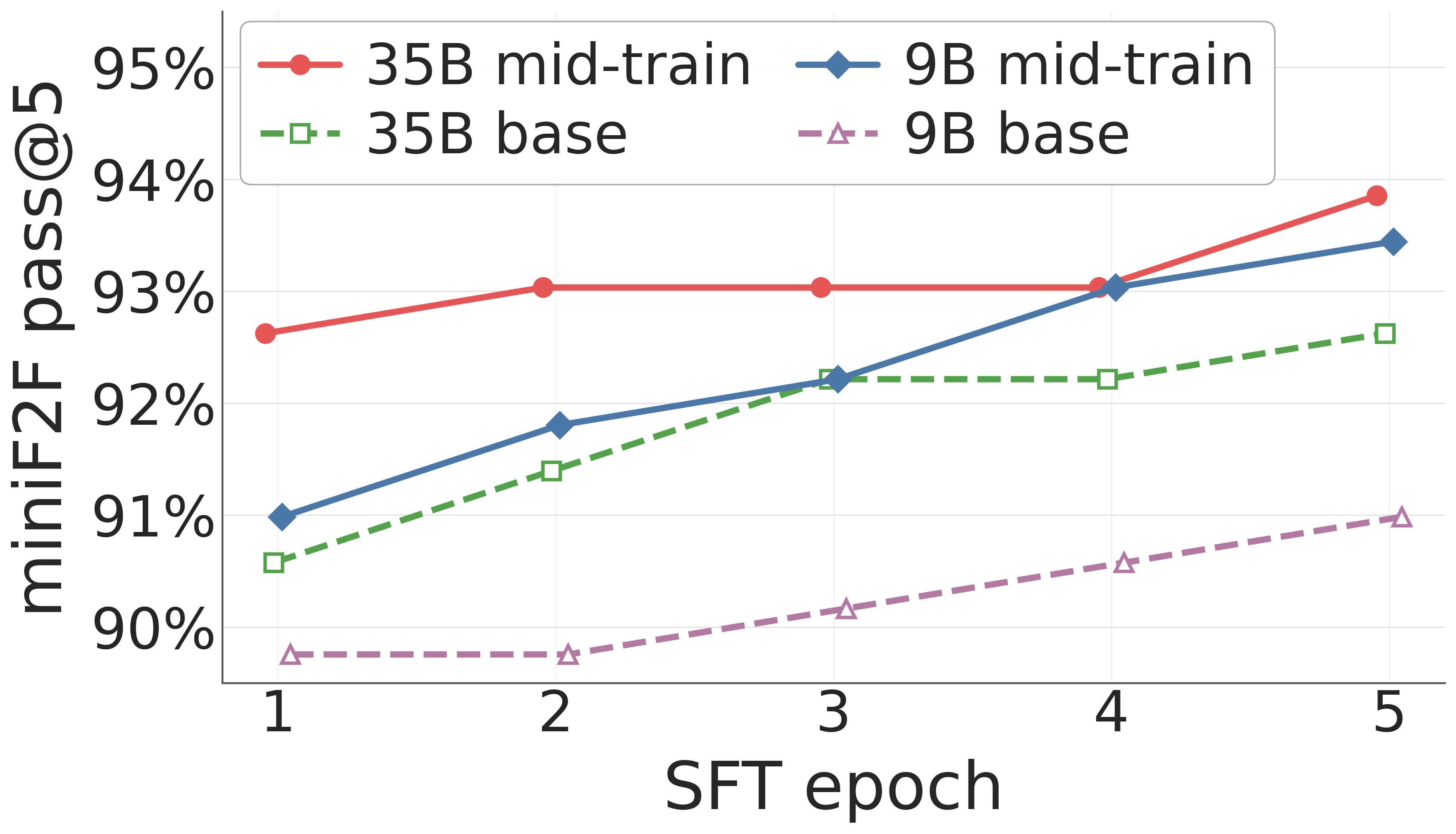}
        \caption{miniF2F $\mathrm{pass}@5$}
        \label{fig:sft_third}
    \end{subfigure}

    \caption{Training dynamics and miniF2F performance of models with and without mid-training.}
    \label{fig:sft_data}
\end{figure}

\textbf{Setup.}  We compare base and mid-trained checkpoints over five epochs of SFT. Each checkpoint is evaluated on miniF2F-test problems using five runs and reports avg@5 and pass@5.

\textbf{Mid-training bridges the gap to agentic learning.}
Figure~\ref{fig:sft_loss} shows that mid-trained checkpoints start SFT with lower loss and remain better aligned with the SFT distribution throughout training. As shown in Figure~\ref{fig:sft_performance} and Figure~\ref{fig:sft_third}, after one epoch, mid-training improves avg@5 by 5.33 and 5.08 points for the 9B and 35B models, respectively, and already matches the base model's epoch-5 pass@5. These gains persist through epoch 5, indicating faster adaptation and a better initialization for downstream SFT.

\textbf{Scaling with model size.}
The 35B-A3B MoE model generally outperforms the 9B model, although the gap is relatively small on MiniF2F, possibly due to limited headroom on this benchmark. As shown in Table~\ref{tab:main_results}, the 35B-A3B midtrain model consistently outperforms the 9B midtrain model across all benchmarks. Notably, the 9B midtrain model surpasses the 35B-A3B base model on both MiniF2F metrics despite having a higher final SFT loss, suggesting that the downstream gains potentially come from mid-training.

\subsection{Analysis of Agent Behavior}

\begin{figure}[htbp]
    \centering

    \begin{subfigure}{0.24\linewidth}
        \centering
        \includegraphics[width=\linewidth]{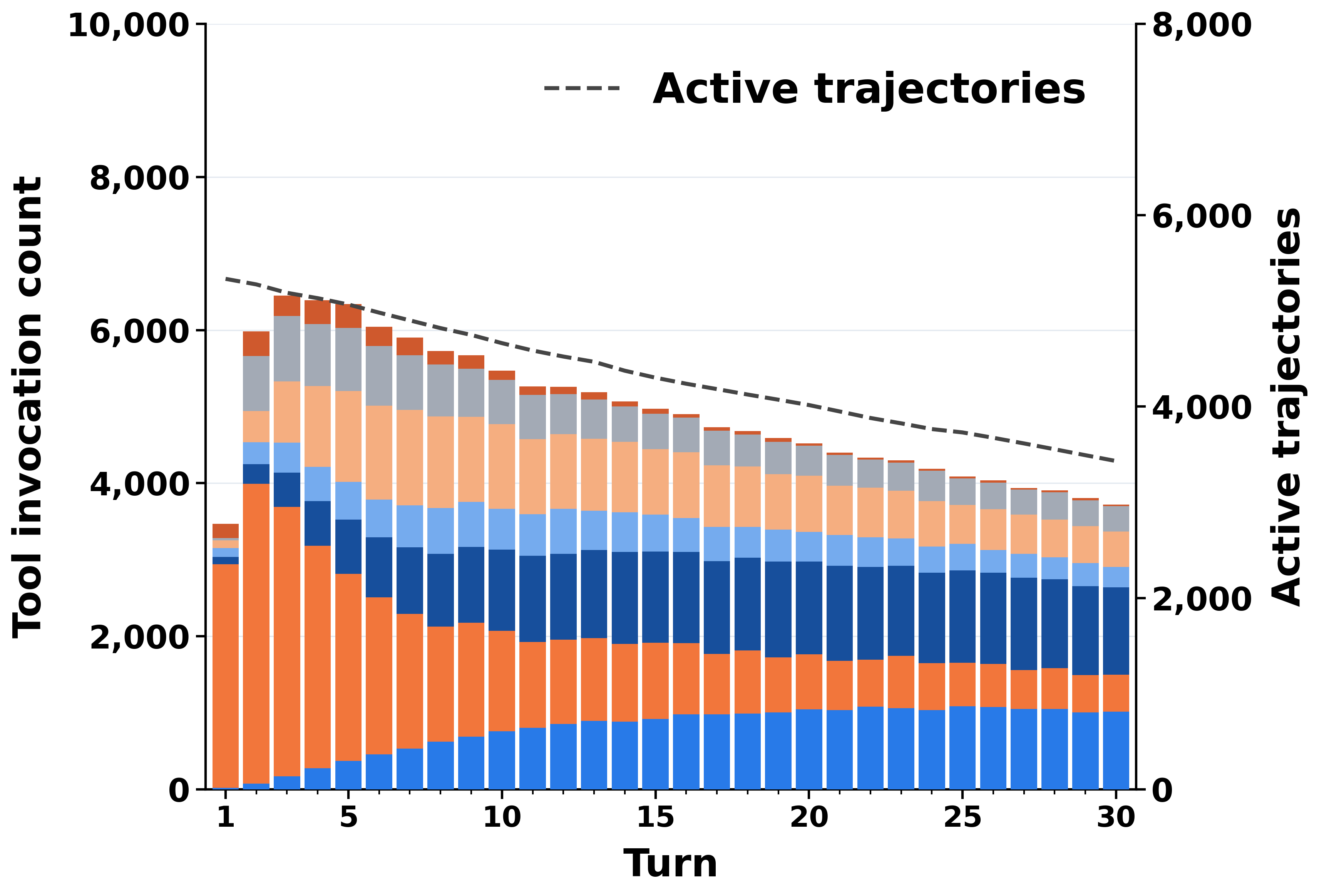}
        \caption{9B Base}
        \label{fig:tool_call_1}
    \end{subfigure}
    \hfill
    \begin{subfigure}{0.24\linewidth}
        \centering
        \includegraphics[width=\linewidth]{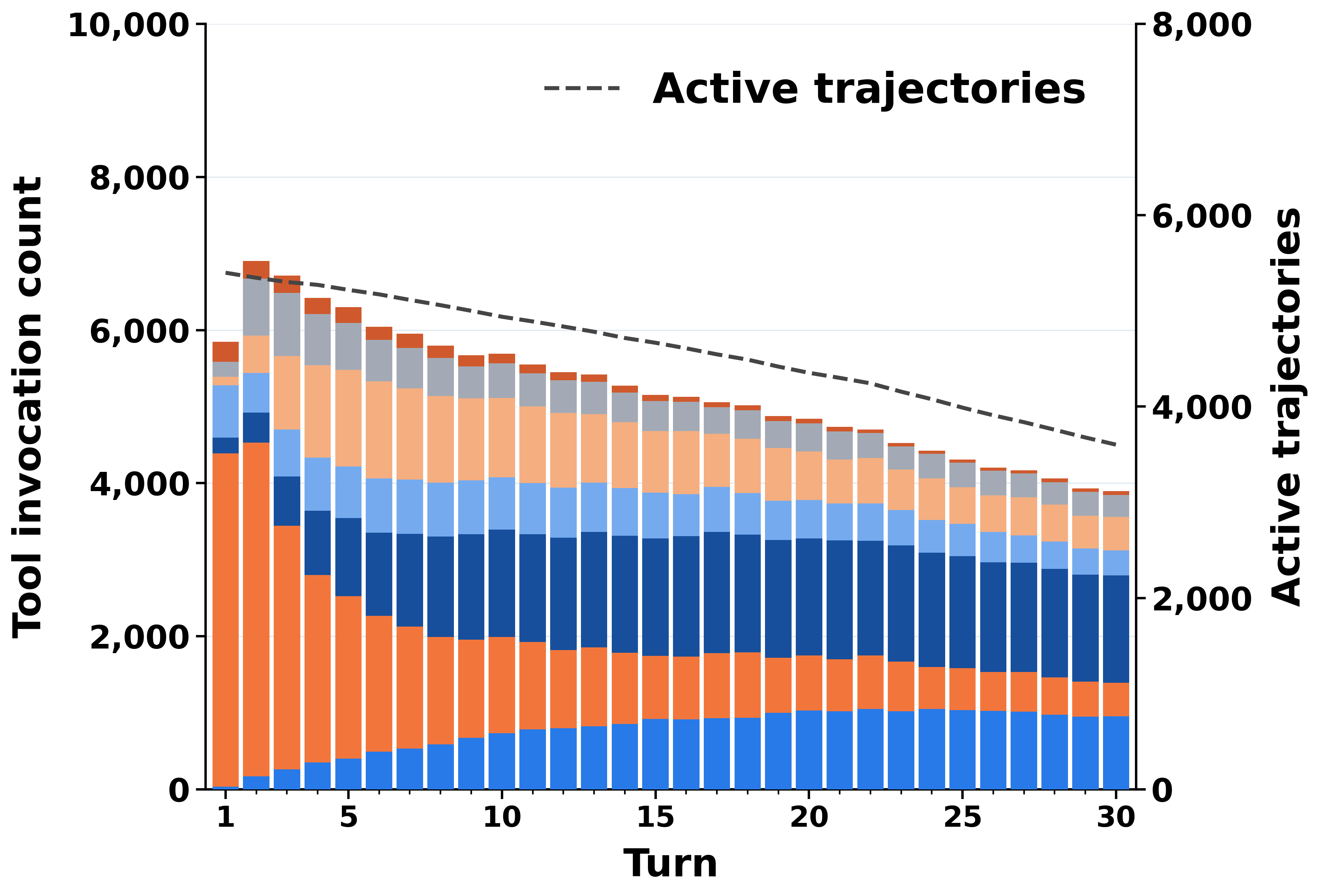}
        \caption{9B Midtrain}
        \label{fig:tool_call_2}
    \end{subfigure}
    \hfill
    \begin{subfigure}{0.24\linewidth}
        \centering
        \includegraphics[width=\linewidth]{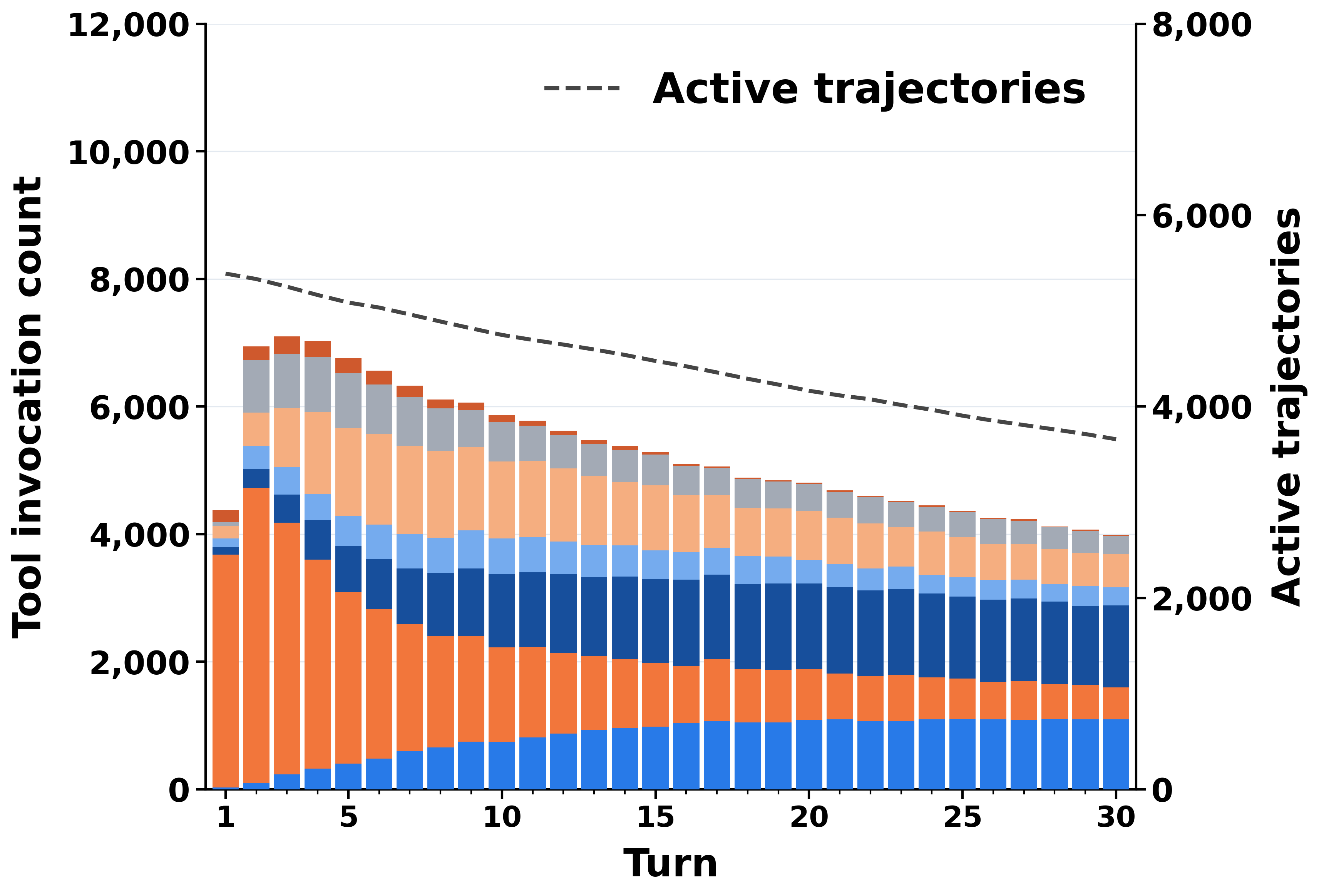}
        \caption{35B-A3B Base}
        \label{fig:tool_call_3}
    \end{subfigure}
    \hfill
    \begin{subfigure}{0.24\linewidth}
        \centering
        \includegraphics[width=\linewidth]{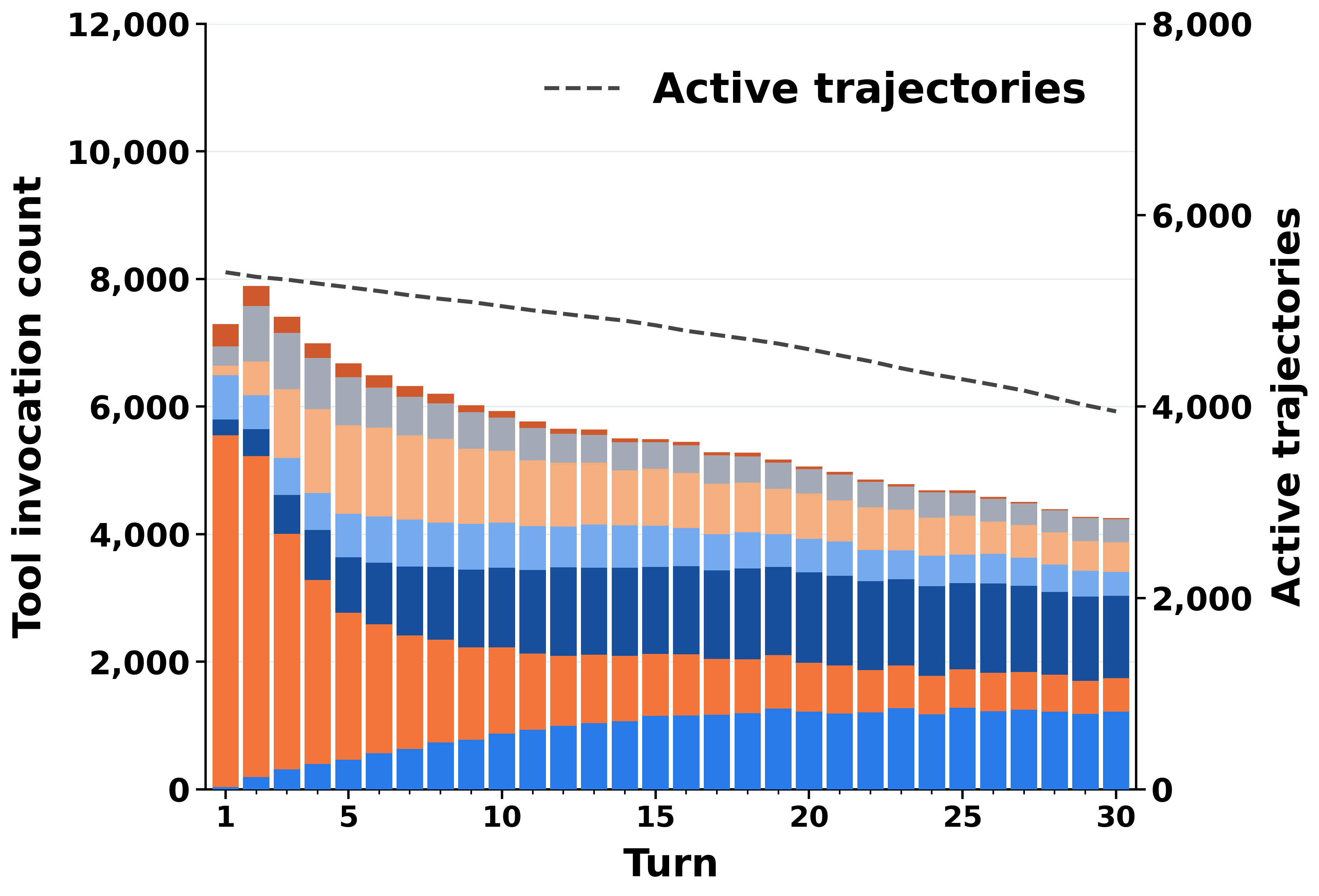}
        \caption{35B-A3B Midtrain}
        \label{fig:tool_call_4}
    \end{subfigure}

     \caption[Agent behaviour.]{%
    Tool call analysis of PutnamBench. Tool colours:
    \protect\tooltag{toolFileEdit}{file\_edit},
    \protect\tooltag{toolLeanRetrieve}{lean\_retrieve},
    \protect\tooltag{toolLeanVerify}{lean\_verify},
    \protect\tooltag{toolFileWrite}{file\_write},
    \protect\tooltag{toolContentSearch}{content\_search},
    \protect\tooltag{toolFileRead}{file\_read}, and
    \protect\tooltag{toolFileSearch}{file\_search}.
    The dashed line indicates the number of active trajectories.
  }
    \label{fig:tool_call}
\end{figure}


\textbf{Common patterns.}
As shown in Figure~\ref{fig:tool_call}, across all four settings, the aggregate tool-use distribution shifts from retrieval towards verification and revision as the interaction progresses. \texttt{lean\_retrieve} accounts for over half of all tool calls during the first five turns, whereas \texttt{lean\_verify} and \texttt{file\_edit} together account for approximately 55--57\% during turns 21--30.  This shift remains visible after normalizing by the number of active trajectories, suggesting a common workflow of gathering relevant information followed by checking and revising candidate proofs. The declining bar heights also reflect the shrinking population of active trajectories. For a relatively easier benchmark such as MiniF2F, we provide a separate analysis in Appendix~\ref{tool_call_miniF2F}, which reveals distinct behavioral patterns.

\begin{wrapfigure}{r}{0.48\linewidth}
    \centering

    \begin{subfigure}[t]{0.48\linewidth}
        \centering
        \includegraphics[width=\linewidth]
            {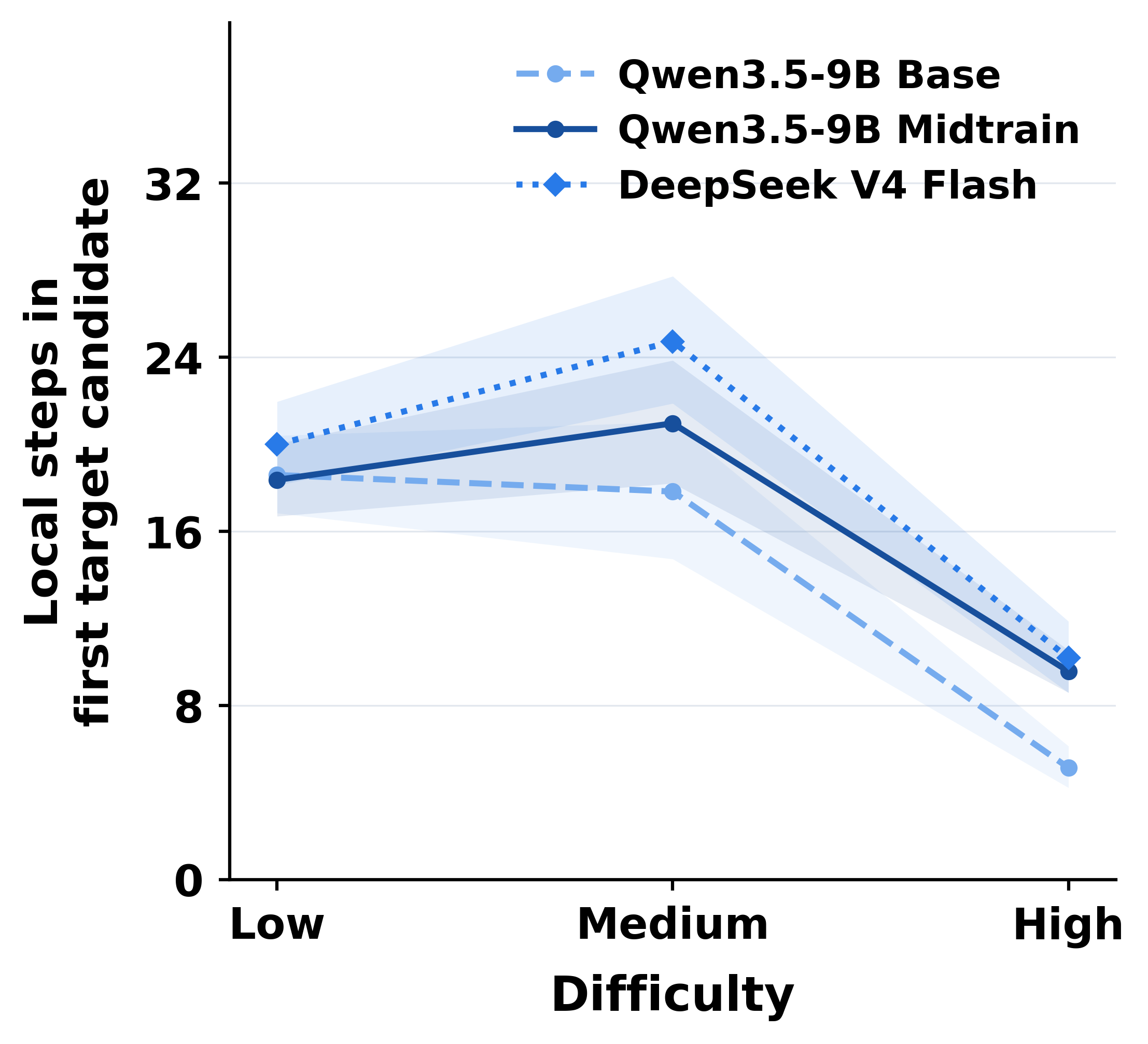}
        \caption{Decomposition}
        \label{fig:decomposition_steps}
    \end{subfigure}
    \hfill
    \begin{subfigure}[t]{0.48\linewidth}
        \centering
        \includegraphics[width=\linewidth]
            {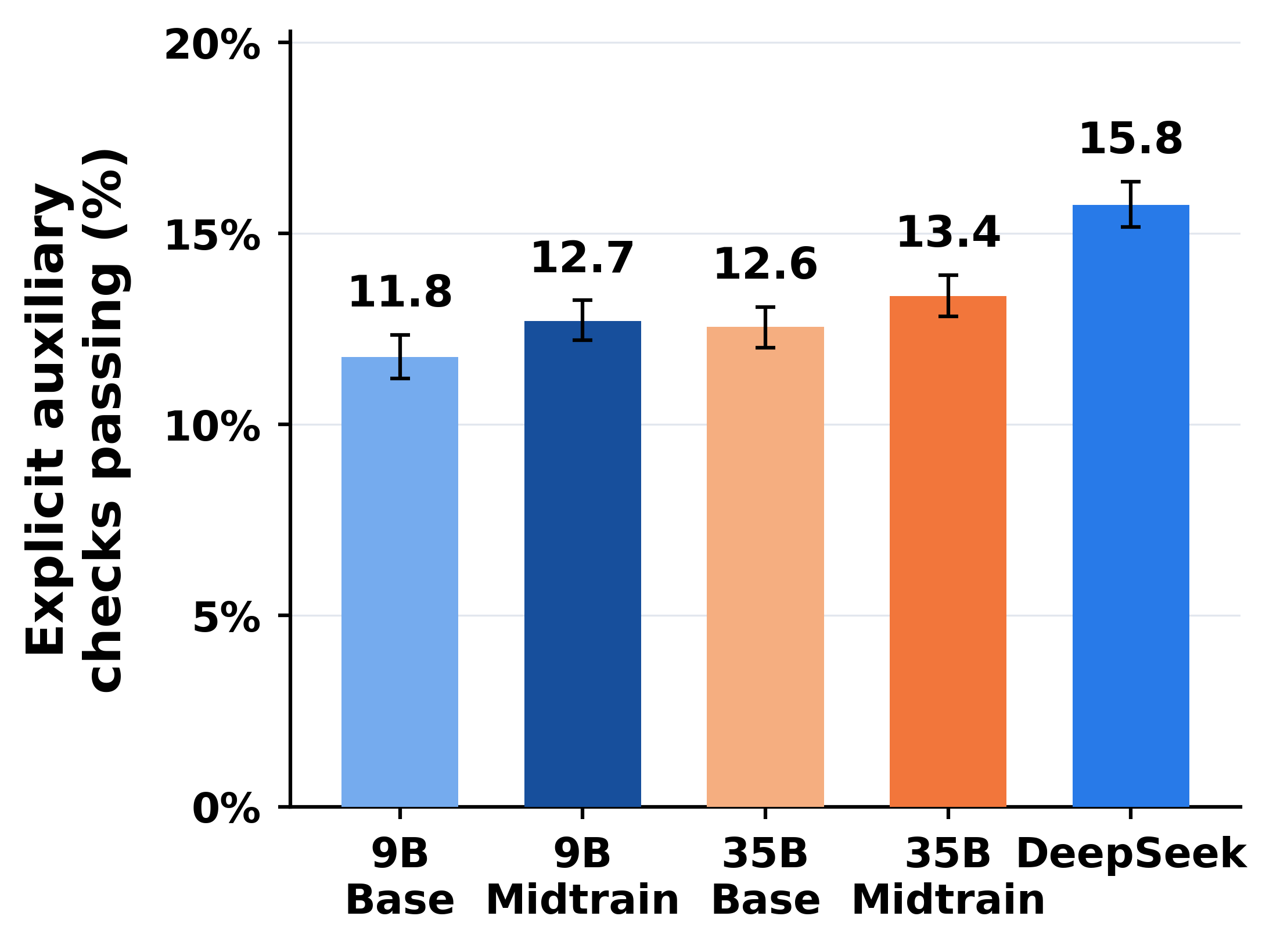}
        \caption{Self-check}
        \label{fig:auxiliary_checks}
    \end{subfigure}

    \caption{Behavior comparison.}
    \label{fig:comparison}
\end{wrapfigure}

\textbf{Midtraining changes model behaviour.}
As shown in Figure~\ref{fig:tool_call}, midtrained models exhibit stronger tool engagement at the beginning of an interaction and increase the share of \texttt{file\_write} calls, particularly in the early turns, indicating earlier file construction. In Figure~\ref{fig:comparison},
explicit auxiliary-check passing improves consistently after midtraining, from 11.8\% to 12.7\% at 9B and from 12.6\% to 13.4\% at 35B.
Moreover, mid-training increases the number of local steps in the first target-candidate proof on medium- and high-difficulty problems, while having little effect on easy problems.   These results suggest that mid-training promotes more decomposition and verification as problem difficulty increases, with the decomposition behavior on harder problems becoming closer to that of DeepSeek-V4-Flash-0731.




\subsection{Scaling with Data Volume}

\begin{wrapfigure}[11]{r}{0.42\linewidth}
    \vspace{-1.8\baselineskip}
    \centering
    \includegraphics[width=\linewidth]
        {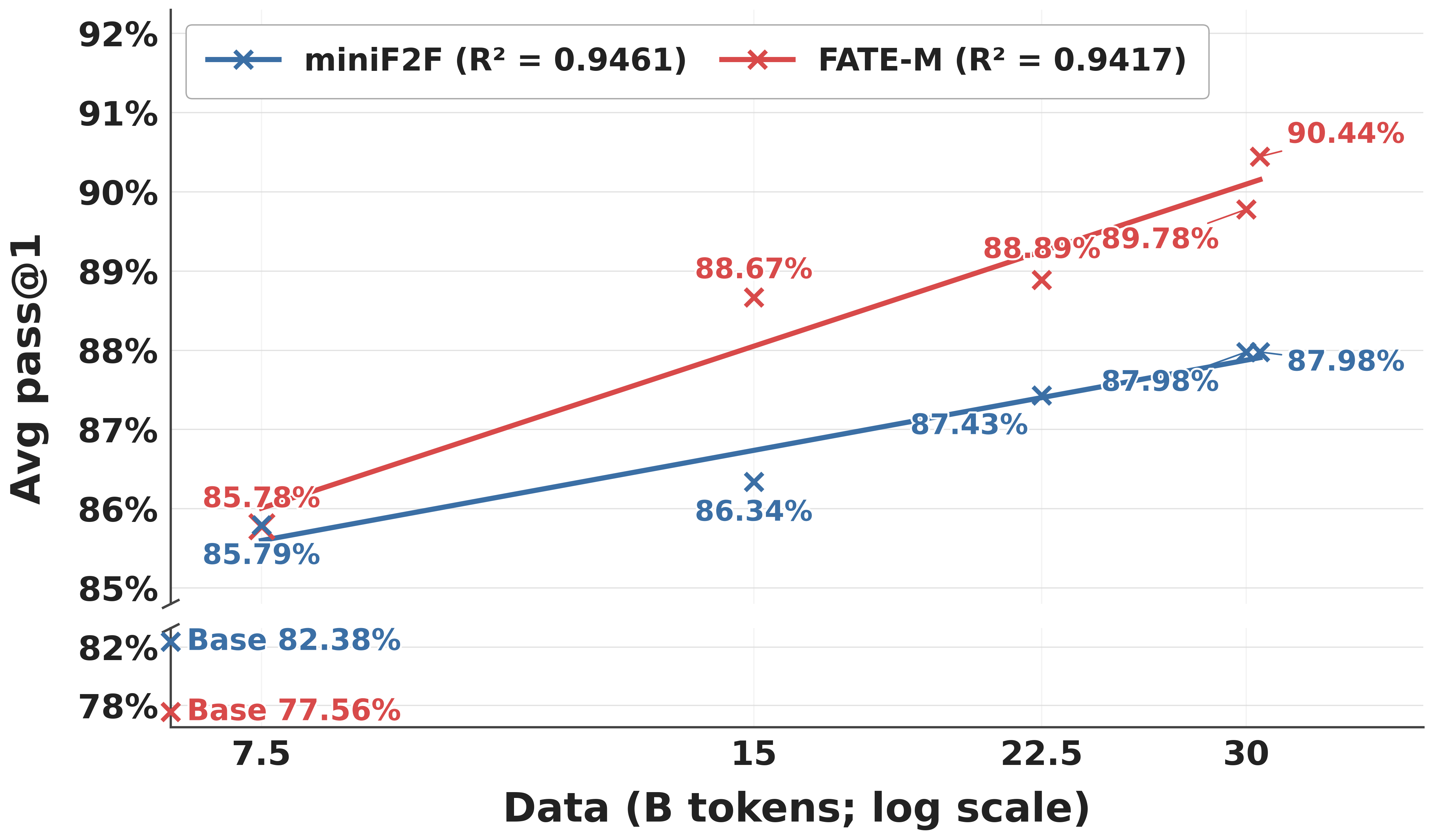}
    \caption{Scaling law of agent-native mid-training.}
    \label{fig:data_scaling}
    \vspace{-1.8\baselineskip}
\end{wrapfigure}

\textbf{Setup.} We evaluate mid-training checkpoints of Qwen3.5-9B at 7.5B, 15B, 22.5B, 30B, and 33.7B tokens. We apply the same two-epoch SFT procedure to each checkpoint and the base model. We measure performance by average pass@1 across three runs.

\textbf{Analysis.} Agentic theorem-proving performance exhibits an approximately log-linear trend with mid- training data scale. Much of the improvement is achieved within the first 15B tokens, yielding gains of 3.96 and 11.11 percentage points over the base model on miniF2F and FATE-M, respectively. Further mid-training generally improves performance, although gains shrink at larger data scales. These results suggest that targeted mid-training can effectively develop agentic theorem proving capabilities.

\subsection{Ablation on daVinci-Prover}

\textbf{Setup.} We explore the contribution of each data component on the Qwen3.5-9B models. 

\textbf{Analysis.} As shown in Table~\ref{tab:data_component_ablation}, the ablation highlights the importance of \emph{agent-native} supervision beyond generic theorem-proving skill acquisition. Despite using only 8.4B tokens, $\mathcal{D}_{agent}$ produces larger gains than the 25.3B-token $\mathcal{D}_{skill}$ on MiniF2F and Fate-M. A likely explanation is distribution alignment: $\mathcal{D}_{agent}$ exposes the model to the same interaction primitives encountered at inference time, thereby bridging the gap between static next-token training and interactive proving. In contrast, $\mathcal{D}_{skill}$ primarily strengthens reusable reasoning priors such as decomposition, search, diagnosis, and self-verification. Importantly, these two forms of supervision are complementary rather than substitutable. Combining them consistently outperforms either component alone, indicating that agentic proving benefits from both knowing how to reason about a proof and knowing how to realize that reasoning through environment interaction.

\begin{table}[t]
    \centering

    \begin{minipage}[t]{0.42\linewidth}
        \centering
        \caption{Data component ablation.}
        \label{tab:data_component_ablation}

        \resizebox{\linewidth}{!}{
            \begin{tabular}{lcccc}
                \toprule
                Component & Tokens & MiniF2F & Fate-M & ProofNet \\
                \midrule
                Qwen3.5-9B Base
                & -- & 87.4 & 88.4 & 42.9 \\
                $\mathcal{D}_{skill}$
                & 25.3B & 87.8 & 89.8 & 43.6 \\
                $\mathcal{D}_{agent}$
                & 8.4B & 88.8 & 92.2 & 43.5 \\
                $\mathcal{D}_{skill} + \mathcal{D}_{agent}$
                & 33.7B & \textbf{90.1} & \textbf{93.3} & \textbf{44.5} \\
                \bottomrule
            \end{tabular}
        }
    \end{minipage}
    \hfill
    \begin{minipage}[t]{0.55\linewidth}
        \centering
        \caption{
        Comparison with existing corpora.
        }
        \label{tab:comparison}

        \resizebox{\linewidth}{!}{
            \begin{tabular}{lccc}
                \toprule
                Data component & MiniF2F & Fate-M & ProofNet \\
                \midrule
                Qwen3-8B-Base + SFT
                & 73.4 & 68.0 & 28.4 \\
                Oprover-8B-Base + SFT
                & 80.3 & 73.1 & 30.3 \\
                Qwen3-Base + Midtrain (ours) + SFT
                & \textbf{85.3} & \textbf{85.3} & \textbf{40.4} \\
                \bottomrule
            \end{tabular}
        }
    \end{minipage}

\end{table}

\subsection{Comparison with Existing Corpora}
\label{corpora_comparison}

\textbf{Setup.}
We perform mid-training on Qwen3-8B-Base using 33B tokens, followed by agentic SFT for 5 epochs. We compare the resulting model against Qwen3-8B-Base without mid-training and Oprover-8B-Base~\citep{ma2026oprover}, which is trained with 65B mid-training tokens. For all base models, we use YaRN~\citep{peng2024yarn} to extend the context length from 32K to 256K.

\textbf{Analysis.}
As shown in Table~\ref{tab:comparison}, our mid-training substantially improves performance over Qwen3-8B-Base across all three benchmarks, confirming that the gains cannot be attributed solely to the subsequent agentic SFT. Moreover, our model consistently outperforms Oprover-8B-Base, with particularly large margins on Fate-M and ProofNet, despite using only 33B mid-training tokens compared with Oprover's 65B. This suggests that the effectiveness of mid-training depends not only on data scale, but critically on its composition: our combination of theorem-proving skill priors and agent-native trajectories provides more targeted supervision for downstream interactive proving and transfers effectively across diverse formal reasoning problems.

\subsection{Memory Compact}

\begin{wraptable}{r}{0.5\textwidth}
  \centering
    \caption{Effect of memory compaction on theorem-proving performance. Performance is evaluated using pass@1, averaged over 3 runs.}
  \label{tab:memory_compact}
  \resizebox{\linewidth}{!}{
  \begin{tabular}{lcccc}
    \toprule
    Model & MiniF2F & FATE-M & FATE-H   \\
    \midrule
    \multicolumn{4}{c}{\textit{w/o memory compact}} \\ 
     Qwen3.5-midtrain-9B  
      & 90.1  & 93.3  & 16.7  \\
    Qwen3.5-midtrain-35B
      & 90.1  & 93.7 &  16.3   \\ \midrule
    \multicolumn{4}{c}{\textit{w/ memory compact}} \\ 
     Qwen3.5-midtrain-9B  
      & 97.1 & 96.7 & 28.0   \\
    Qwen3.5-midtrain-35B
      & 97.1  & 97.3 & 42.0   \\
    \bottomrule
  \end{tabular}
  }
\end{wraptable}

\textbf{Setup.} To support long-horizon difficult proofs, we use a simple, model-free trajectory compression method. Once the context length reaches 200K tokens, we trigger compression by replacing older interactions with deterministically extracted tool evidence, while preserving the original theorem, current Lean proof, and recent interactions. We allow up to 8 compactions, and other settings remain unchanged.

\textbf{Analysis.}
Memory compaction consistently improves performance, with the largest gains on the more challenging FATE-H benchmark. The improvement is especially pronounced for the 35B model, suggesting that effective context management is important for long-horizon proving and can better expose the capacity of larger models.

\section{Conculusion}

We introduced daVinci-Prover, combining agent-native mid-training with verifier-grounded supervised fine-tuning for interactive theorem proving based on a single harness. Experiments demonstrate consistent improvements over SFT-only baselines, while ablations support the complementary roles of mathematical skill priors and environment-native trajectories. These findings highlight agent-native mid-training as an effective foundation for developing stronger theorem-proving agents.

\subsection*{AI USE Statement}
In this work, we used generative AI tools for paper polishing and coding assistance. All AI-assisted content was carefully reviewed by the authors. For example, we manually checked AI-polished text, and AI-generated code was independently verified and tested for correctness by two authors. We take full responsibility for the final content of this work, including all text, claims, and artifacts produced with the assistance of generative AI.

\subsection*{REPRODUCIBILITY STATEMENT}
The data we collected, the models we trained, the harness we built, and the corresponding code and training infrastructure will be fully open-sourced for reproducibility after the anonymous review period ends.

\subsection*{Ethics statement}
No ethics statement is required here.

\bibliography{iclr2027_conference}
\bibliographystyle{iclr2027_conference}

\appendix

\section*{Limitation}



During the mid-training phase, we do not replay any generalization data. However, we believe that incorporating generalization replay is both necessary and important for maintaining and improving the model's generalization capabilities.

\section{Related Work}
\label{sec:re}

\paragraph{Mid-training.}
Recent work views mid-training as a useful bridge between pre-training and post-training. Instead of moving directly from pre-training to SFT or RL, mid-training uses cleaner and more task-relevant data to better prepare the model, often with learning-rate annealing~\citep{zhang2025interplay, tu2025survey}. For example, OctoThinker~\citep{wang2025octothinker} shows that reasoning-focused mid-training can improve both RL efficiency and final performance.

\paragraph{Agentic Proving}

Agentic provers make the interaction loop part of the solution. Prover Agent coordinates informal reasoning, formal proving, auxiliary lemmas, and Lean feedback~\citep{baba2025prover}. OProver retrieves verified proofs, repairs failures, and feeds successful interactions back into training~\citep{ma2026oprover}; LongCat-Flash-Prover trains auto-formalization, sketching, proving, and tool use through agentic reinforcement learning~\citep{wang2026longcat}. Thus, interaction traces can serve as training data rather than only inference-time scaffolding.

\paragraph{Agent-Native Mid-Training}
Software-engineering research offers a complementary training perspective: Kimi-Dev treats structured workflow training as transferable skill priors, while DaVinci-Dev combines contextually and environmentally native trajectories during mid-training~\citep{yang2026kimi,zeng2026davinci}. We also focus on the stage at which agents acquire capabilities. We organize decomposition, search, revision, and environment interaction as complementary mid-training skills, then study whether these priors transfer from whole-proof generation to downstream interactive proving.

\section{In-depth analysis of Agent Behavior}

\begin{figure}[htbp]
    \centering

    \begin{subfigure}{0.32\linewidth}
        \centering
        \includegraphics[width=\linewidth]{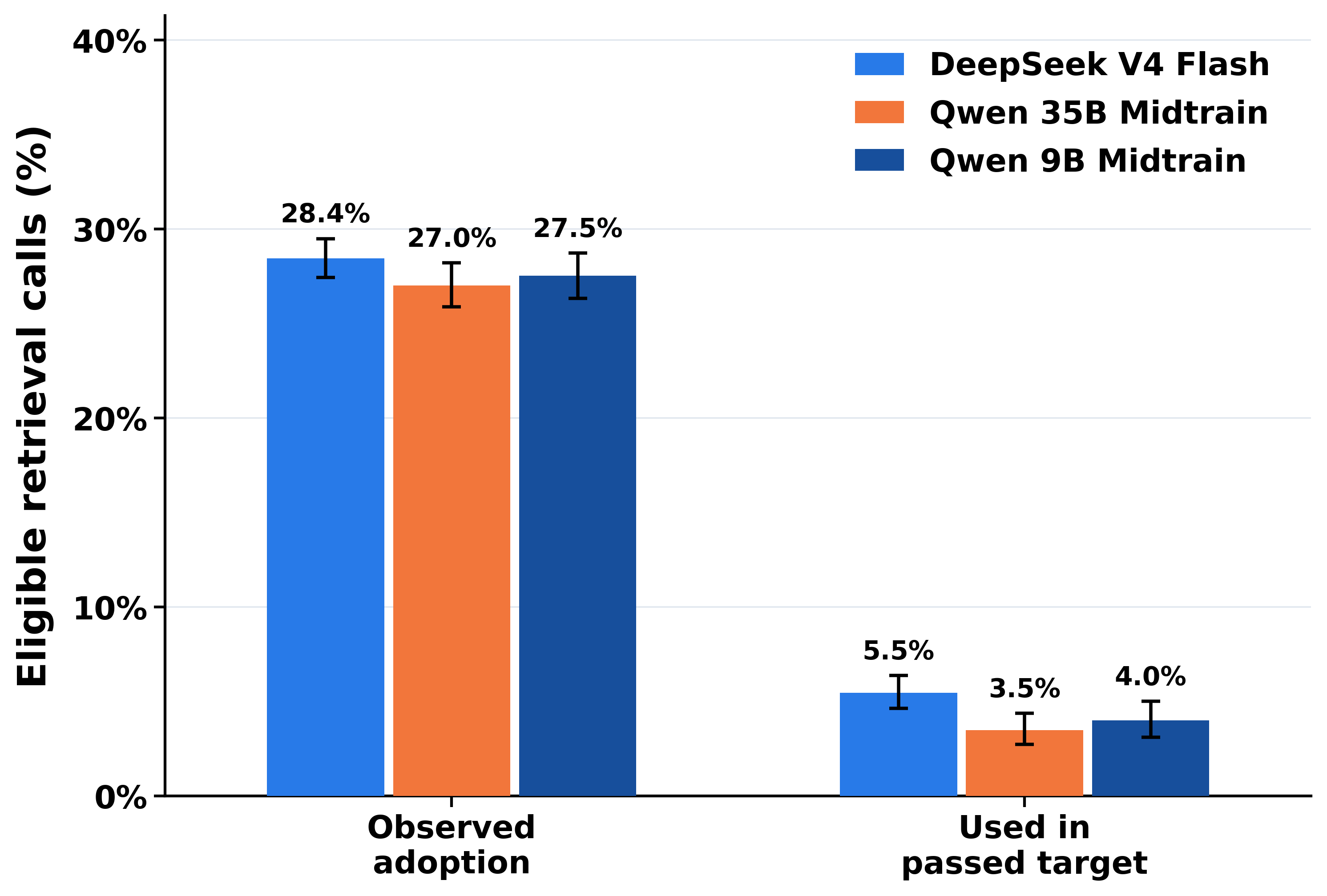}
        \caption{Retrival calls}
        \label{fig:in_depth_of_agent_behavior_1}
    \end{subfigure}
    \hfill
    \begin{subfigure}{0.32\linewidth}
        \centering
        \includegraphics[width=\linewidth]{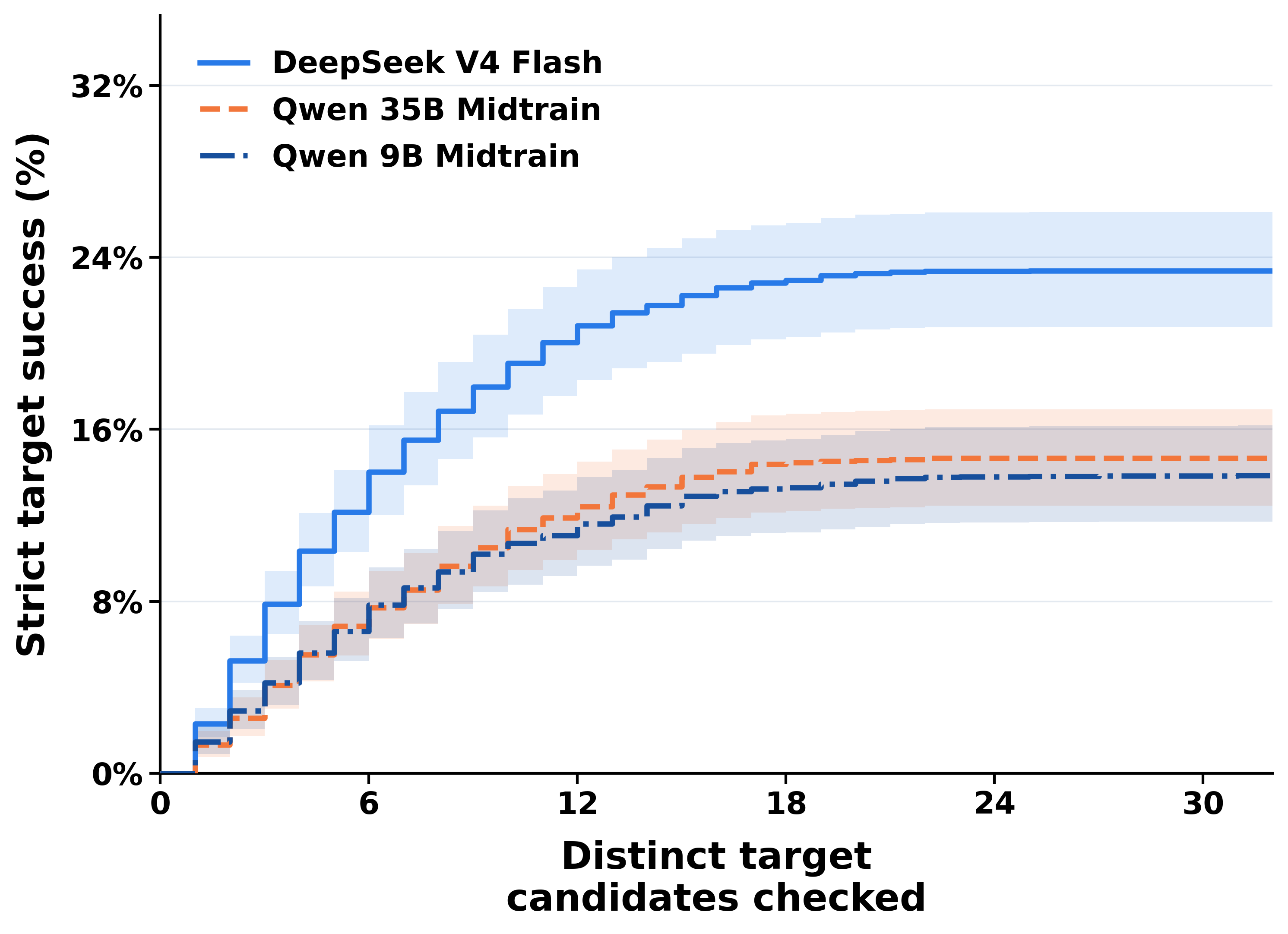}
        \caption{Strict proof success}
        \label{fig:in_depth_of_agent_behavior_2}
    \end{subfigure}
    \hfill
    \begin{subfigure}{0.32\linewidth}
        \centering
        \includegraphics[width=\linewidth]{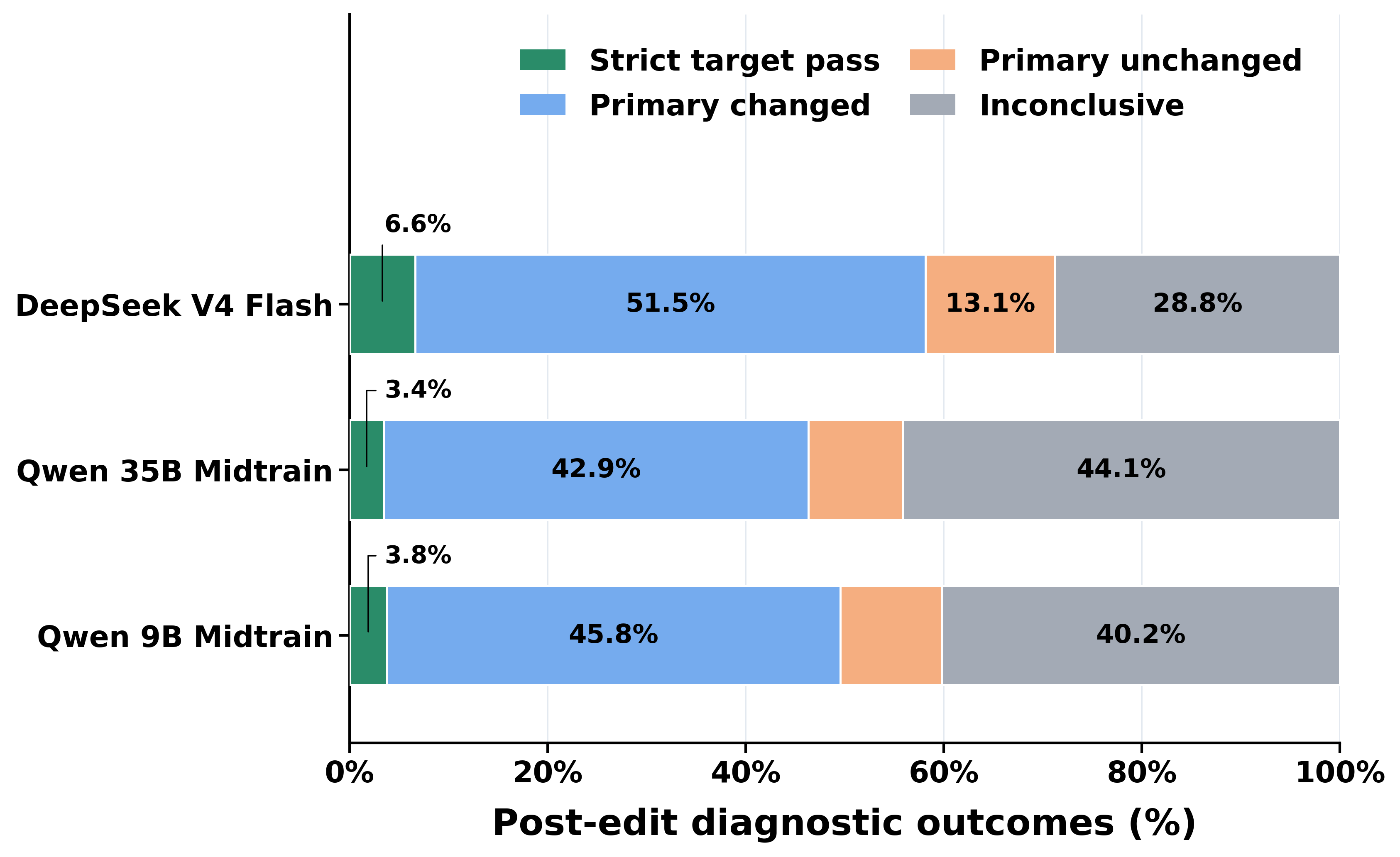}
        \caption{Revision capability}
        \label{fig:in_depth_of_agent_behavior_3}
    \end{subfigure}
    \caption{In-depth analysis of agent trajectories.}
    \label{fig:in_depth_of_agent_behavior}
\end{figure}

We analyze the agent trajectories on the same 672 Putnam problems, using eight runs for DeepSeek-V4-Flash-0731 and eight runs for Qwen3.5-9B midtrain and Qwen3.5-35B-A3B midtrain models.

\textbf{Retrieval utilization.}
DeepSeek-V4-Flash-0731, Qwen3.5-35B-A3B-Midtrain, and Qwen3.5-9B-Midtrain exhibit similar observed retrieval adoption rates of 28.45\%, 27.01\%, and 27.53\%, respectively. The fraction of retrieval calls whose returned declarations appear in strictly verified target proofs is higher for DeepSeek (5.47\%) than for Qwen3.5-35B (3.48\%) and Qwen3.5-9B (4.01\%). Both measures use all successful, nonempty retrieval calls as their denominator. 

\textbf{Proof completion across candidate checks.}
DeepSeek's advantage is already visible within the first few candidate checks. By the fifth distinct target candidate checked, cumulative observed success reaches 12.15\%, compared with 6.85\% and 6.60\% for Qwen 35B and 9B. Over the full recorded trajectories, the corresponding rates reach 23.38\%, 14.66\%, and 13.86\%. This pattern shows higher proof completion across the observed candidate generation and revision process. The horizontal axis counts distinct checked target candidates, including drafts, rather than interaction turns.

\textbf{Outcomes of iterative edits.}
Among content-changing edits following a failed check of the same target file with a primary diagnostic, 6.60\% of DeepSeek's edits immediately produce a strict target pass, compared with 3.43\% and 3.75\% for Qwen 35B and 9B. DeepSeek therefore completes the target more frequently immediately after eligible edits in these logs.

\subsection{Tool call analysis of MiniF2F}
\label{tool_call_miniF2F}
\begin{figure}[htbp]
    \centering

    \begin{subfigure}{0.24\linewidth}
        \centering
        \includegraphics[width=\linewidth]{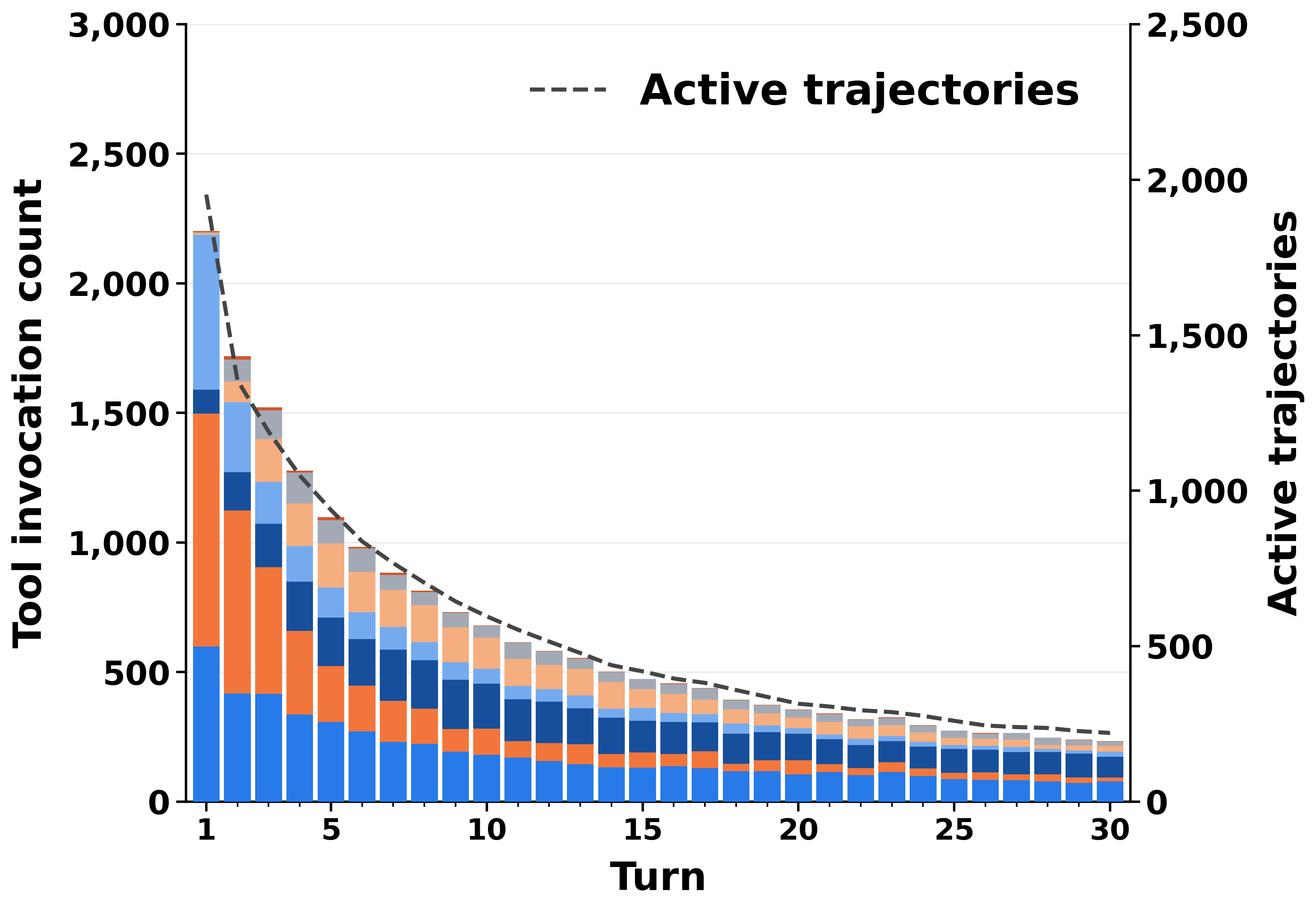}
        \caption{9B Base}
        \label{fig:tool_call_minif2f_1}
    \end{subfigure}
    \hfill
    \begin{subfigure}{0.24\linewidth}
        \centering
        \includegraphics[width=\linewidth]{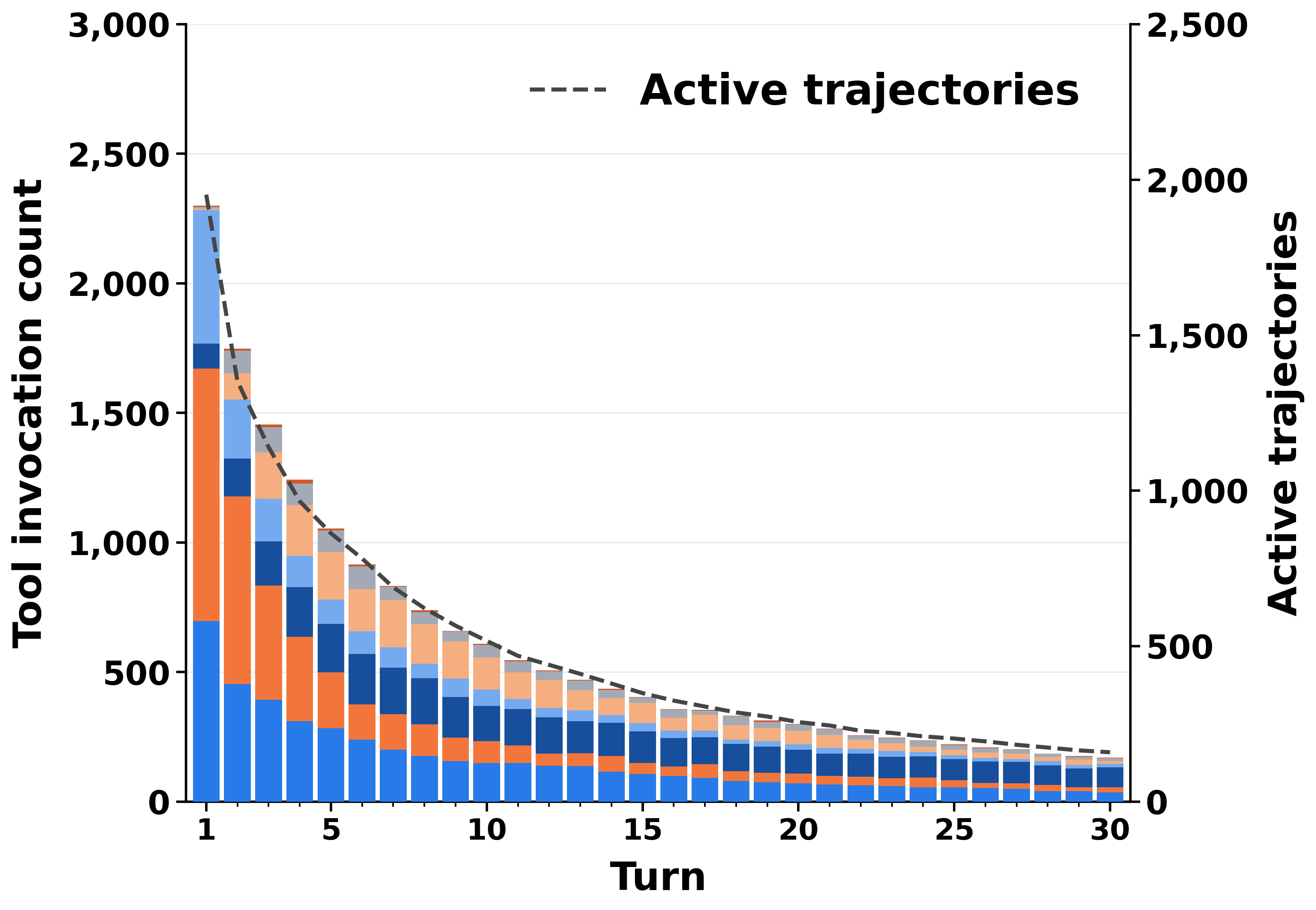}
        \caption{9B Midtrain}
        \label{fig:tool_call_minif2f_2}
    \end{subfigure}
    \hfill
    \begin{subfigure}{0.24\linewidth}
        \centering
        \includegraphics[width=\linewidth]{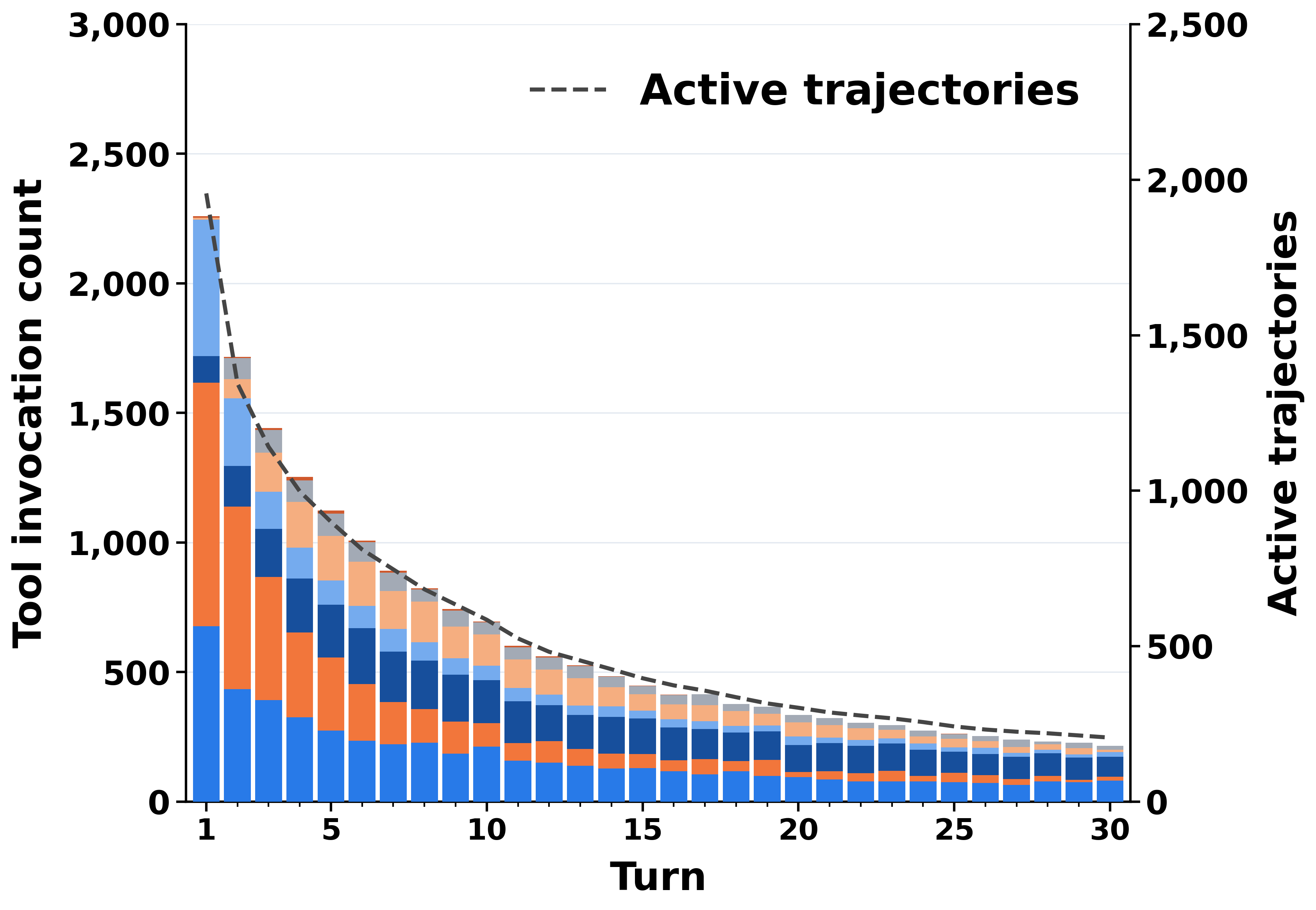}
        \caption{35B-A3B Base}
        \label{fig:tool_call_minif2f_3}
    \end{subfigure}
    \hfill
    \begin{subfigure}{0.24\linewidth}
        \centering
        \includegraphics[width=\linewidth]{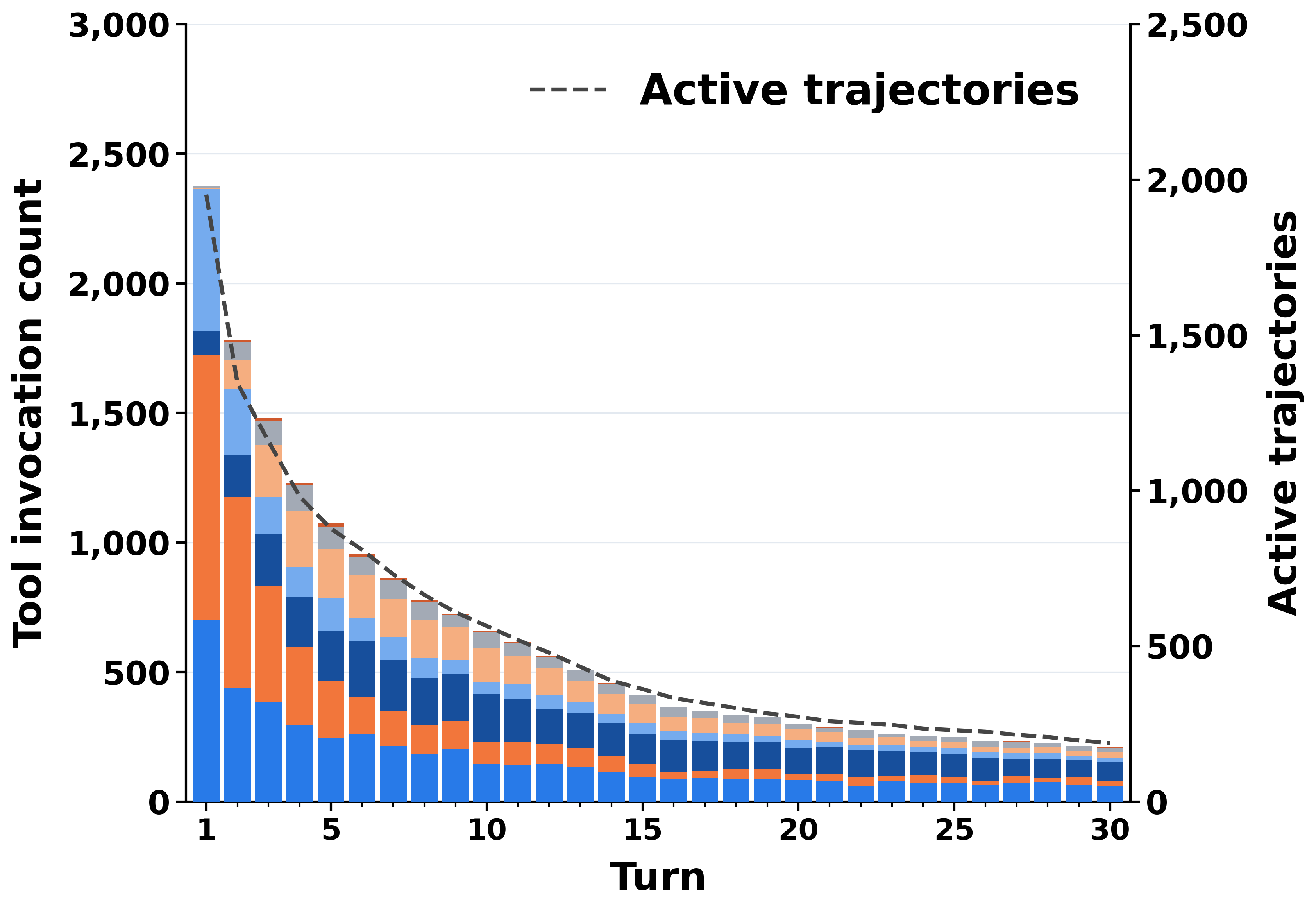}
        \caption{35B-A3B Midtrain}
        \label{fig:tool_call_minif2f_4}
    \end{subfigure}

     \caption[Agent behaviour.]{%
    Tool call analysis. Tool colours:
    \protect\tooltag{toolFileEdit}{file\_edit},
    \protect\tooltag{toolLeanRetrieve}{lean\_retrieve},
    \protect\tooltag{toolLeanVerify}{lean\_verify},
    \protect\tooltag{toolFileWrite}{file\_write},
    \protect\tooltag{toolContentSearch}{content\_search},
    \protect\tooltag{toolFileRead}{file\_read}, and
    \protect\tooltag{toolFileSearch}{file\_search}.
    The dashed line indicates the number of active trajectories.
  }
    \label{fig:tool_call_miniF2F}
\end{figure}

Compared with Putnam, miniF2F shows shorter tool-use trajectories and earlier proof writing and editing. Only about 8–11\% of trajectories remain active at turn 30 on miniF2F, compared with 64–73\% on Putnam. Putnam also places greater emphasis on early retrieval, followed by sustained editing and verification.

Across the four miniF2F models, the overall tool-use pattern is similar. Midtrained models make fewer total tool calls while recording slightly more successful tasks, with a larger reduction at 9B (8.75\%) than at 35B (2.65\%). Their late-stage active trajectories also decline more quickly. On Putnam, however, Midtrained models sustain longer interactions and make more calls. This suggests that Midtraining changes interaction duration differently across benchmarks.

\subsection{Decomposition}

\begin{figure}[htbp]
    \centering

    \begin{subfigure}{0.48\linewidth}
        \centering
        \includegraphics[width=\linewidth]
        {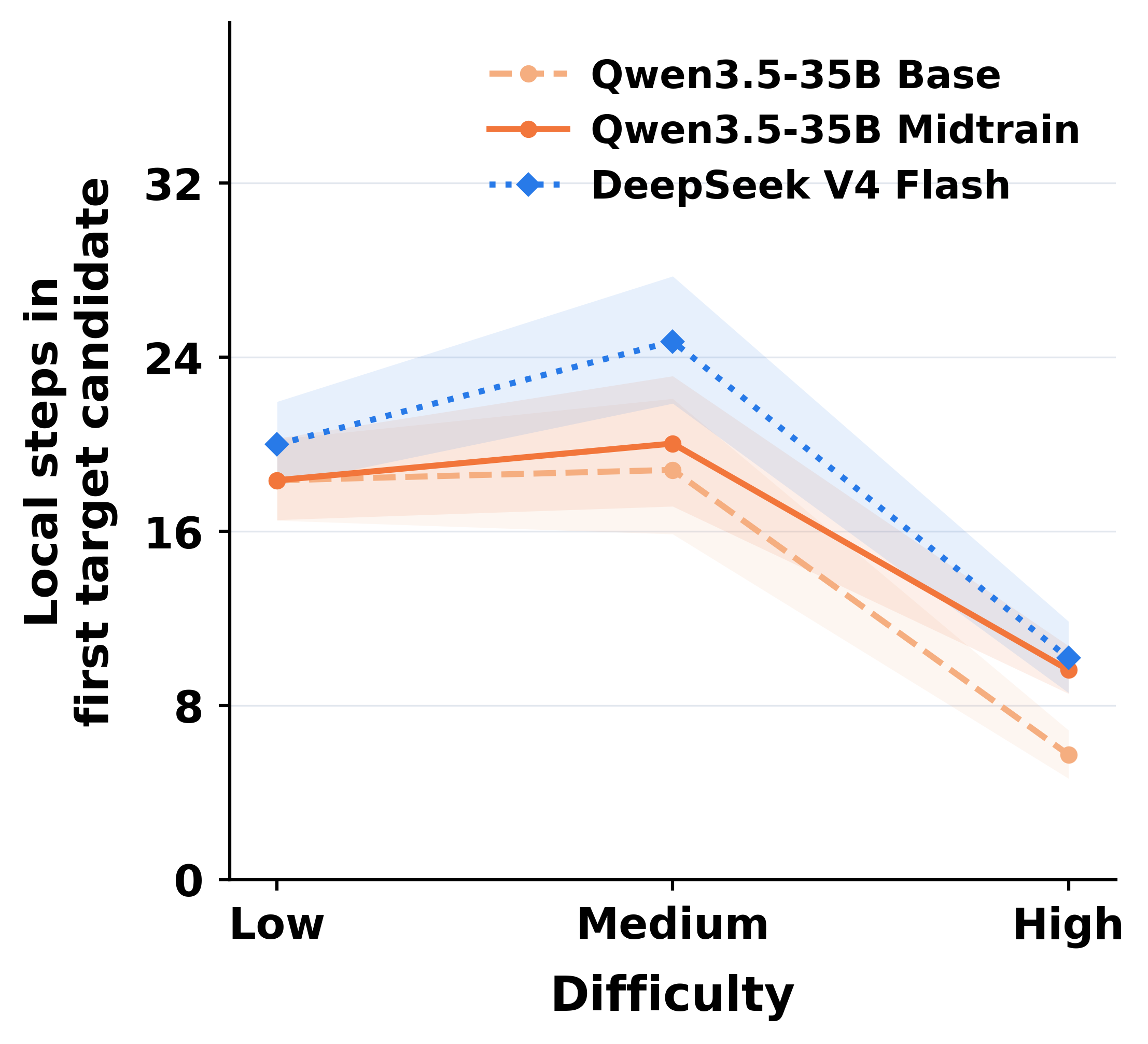}
        \caption{Qwen3.5-35B-A3B, Local steps in proof}
        \label{fig:subfig1}
    \end{subfigure}
    \hfill
    \begin{subfigure}{0.48\linewidth}
        \centering
        \includegraphics[width=\linewidth]
        {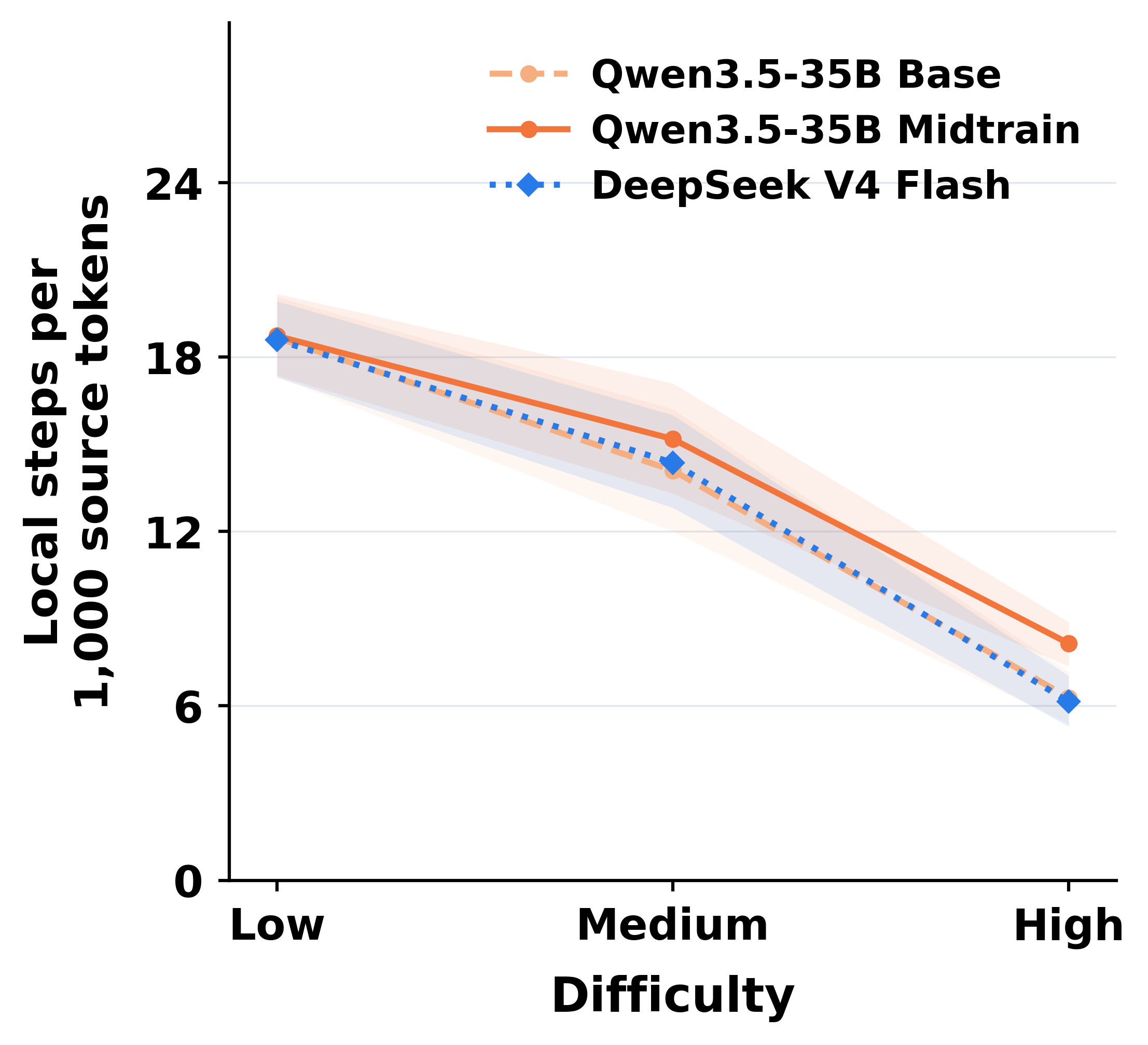}
        \caption{Qwen3.5-35B-A3B, Local steps in 1k tokens}
        \label{fig:subfig2}
    \end{subfigure}

    \par\medskip

    \begin{subfigure}{0.48\linewidth}
        \centering
        \includegraphics[width=\linewidth]
        {figures/putnam_qwen9b_decomposition_steps_by_difficulty.png}
        \caption{Qwen3.5-9B, Local steps in proof}
        \label{fig:subfig3}
    \end{subfigure}
    \hfill
    \begin{subfigure}{0.48\linewidth}
        \centering
        \includegraphics[width=\linewidth]
        {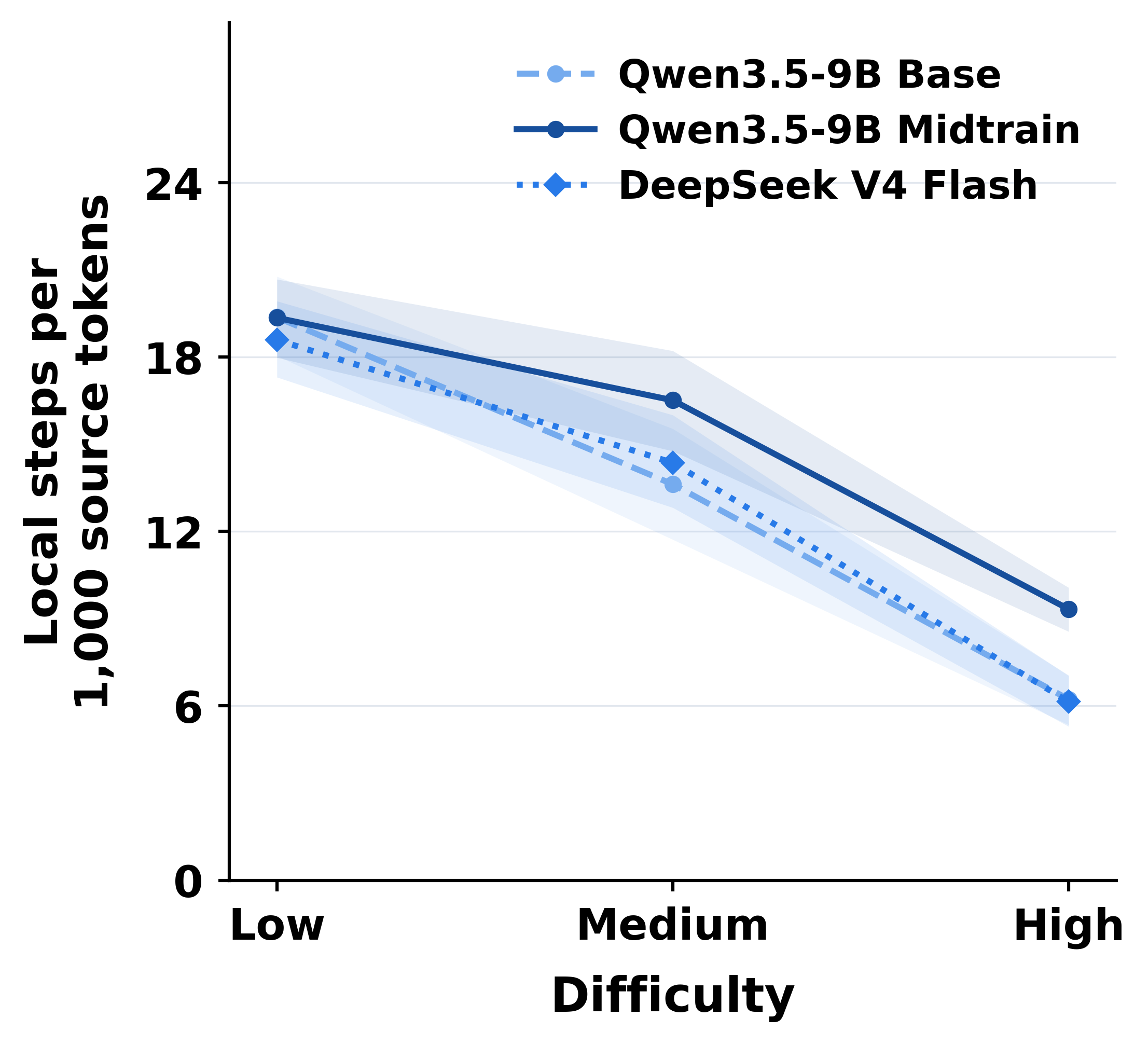}
        \caption{Qwen3.5-9B, Local steps in 1k tokens}
        \label{fig:subfig4}
    \end{subfigure}

    \caption{Proof decomposition analysis.}
    \label{fig:problem_decom}
\end{figure}

\textbf{Effect of midtraining on proof decomposition.}
Midtrained models produce more explicit intermediate steps in their first checked target proof, particularly on harder problems. In the high-difficulty group, defined as problems unsolved by the DeepSeek reference across eight attempts, the average number of local \texttt{have}/\texttt{suffices} steps increases from 5.13 to 9.57 at 9B and from 5.73 to 9.63 at 35B. The corresponding step densities also increase, from 6.22 to 9.33 and from 6.28 to 8.15 per 1,000 lexical source tokens, respectively. Thus, these differences are not explained solely by longer code. By contrast, Base and Midtrain models show similar step counts in the low difficulty group. This pattern suggests that midtraining is associated with more explicit decomposition when constructing initial proofs for challenging problems.

\textbf{Comparison with DeepSeek.} DeepSeek produces more local steps in absolute terms, especially in the medium-difficulty group, but its step density is lower than that of the Midtrained Qwen models. In the high-difficulty group, its average step count is 10.17, close to the approximately ten steps produced by both Midtrained Qwen models. Therefore, DeepSeek's higher overall success rate cannot be attributed simply to producing more intermediate steps.

\section{Decontamination}
\label{decontamination}

To ensure a fair evaluation, we perform benchmark-level
decontamination against all evaluation sets listed in
Table~\ref{tab:benchmark-overlap}. For Lean data, we extract the theorem
signature by removing nested comments and proof bodies, normalize
Unicode characters, tokenize the remaining statement, and anonymize
the declaration name. We then perform exact matching against all
available benchmark variants. To avoid false positives caused by short
or generic statements, a token-level formal match is automatically
confirmed only if the anonymized signature contains at least 20 tokens
and 80 normalized characters. Exact provenance links and byte-identical
raw statements are also treated as confirmed matches.

For informal data, we apply Unicode and math-aware normalization and
require an exact match of the complete problem statement with at least
12 content tokens. Matches based only on theorem names or approximate
textual similarity are not automatically classified as contamination;
instead, they are reserved for manual review. For each confirmed match,
we remove the entire logical parent example, including all of its
semantic chunks and any replayed copies derived from it.

Before filtering, we identify 29 overlapping problems among the 2,137
evaluation problems (1.36\%). When conservatively counting every token
in each matched parent trajectory, these examples account for at most
11.55M tokens, corresponding to 0.03427\% of the 33.7B-token
mid-training corpus.

\begin{table}[hpt]
    \centering
    \caption{
        Confirmed benchmark overlaps detected in the mid-training
        corpus before decontamination. Approximate and review-only
        matches are excluded.
    }
    \label{tab:benchmark-overlap}
    \begin{tabular}{lrrr}
        \toprule
        \textbf{Benchmark}
        & \textbf{Overlaps}
        & \textbf{Total}
        & \textbf{Rate} \\
        \midrule
        miniF2F                 & 16 & 244 & 6.56\% \\
        MathOlympiadBench       &  9 & 360 & 2.50\% \\
        ProofNet                &  2 & 186 & 1.08\% \\
        DeepSeek-ProverBench    &  1 & 325 & 0.31\% \\
        Putnam                  &  1 & 672 & 0.15\% \\
        FATE-H/M/X              &  0 & 350 & 0.00\% \\
        \midrule
        \textbf{Total}
        & \textbf{29}
        & \textbf{2,137}
        & \textbf{1.36\%} \\
        \bottomrule
    \end{tabular}
\end{table}

For the SFT training data, we also remove any examples that overlap with the evaluation benchmarks.

\section{Evaluation metric}

In Table~\ref{tab:main_results}, to account for inference efficiency and budget, we evaluate each problem using Pass@8:
\begin{equation}
    \operatorname{Pass@8}(i)
    =
    \mathbf{1}\left[
        \sum_{j=1}^{8} \mathrm{success}_{ij} > 0
    \right].
\end{equation}




\section{Tool Schema}
\label{tool_schema}
We use the following tools to expose the agent. There are only 7 tools. For the convenience of future RL expansion, we do not use bash and Python commands to expose the agent.

\begin{table}[hpt]
\centering
\setlength{\tabcolsep}{3pt}
\renewcommand{\arraystretch}{1.5}
\caption{Internal tools exposed to the davici-Prover agent harness.}
\label{tab:davici-agent-tools}
\begin{tabularx}{\textwidth}{@{}
>{\raggedright\arraybackslash}p{0.3\textwidth}
>{\raggedright\arraybackslash}p{0.2\textwidth}
>{\raggedright\arraybackslash}X@{}}
\toprule
\textbf{Tool} & \textbf{Category} & \textbf{Description} \\
\midrule
\path|file_read| & File & Read workspace files or line ranges. \\
\path|file_write| & Edit & Write complete workspace files. \\
\path|file_edit| & Edit & Replace text spans or line ranges. \\
\path|file_search| & Search & Find files by glob patterns. \\
\path|content_search| & Search & Search file contents by text or regex. \\
\path|lean_retrieve| & Retrieval & Retrieve relevant Mathlib declarations. \\
\path|lean_verify| & Verification & Check Lean code or scratch files. \\
\bottomrule
\end{tabularx}
\end{table}

\section{Data Synthesis Costs}
\label{sec:data_synthesis_costs}

\textbf{Teacher deployment.}
We deployed DeepSeek-V4-Flash-0731 on 48 NVIDIA H200 GPUs for approximately 20 days, with the teacher service shared across mid-training and SFT data synthesis. This corresponds to approximately $48 \times 20 \times 24 = 23{,}040$ allocated H200 GPU-hours. This deployment-level budget includes unsuccessful attempts, discarded generations, and idle periods, in addition to generations retained in the final datasets.

\textbf{CPU resources for Lean verification.}
The CPU service supporting Lean verification used approximately eight worker instances, each provisioned with 20 CPU cores and 80\,GB of memory, for roughly the same 20-day period. The estimated aggregate allocation was therefore 160 CPU cores and 640\,GB of memory, corresponding to approximately $8 \times 20 \times 20 \times 24 = 76{,}800$ allocated CPU core-hours. 

\textbf{Public-cloud rental equivalent.}
For a public-cloud comparison, Runpod lists H200 Secure Cloud instance and H200 SXM cluster rates of US\$5.29 and US\$6.35 per GPU-hour, respectively, as of November 9, 2026.\footnote{\url{https://www.runpod.io/pricing}, accessed November 9, 2026.} Applying these rates to the teacher deployment yields a GPU rental-equivalent cost of approximately US\$122{,}000--146{,}000. This is a reference-price estimate rather than an actual expenditure figure, and excludes the additional CPU service for Lean verification and separate storage charges.

\section{Hyperparameters}

\subsection{Mid-Training Details}

We perform full-parameter mid-training using the autoregressive
next-token prediction objective, while retaining the one-layer multi-token
prediction (MTP) objective with a loss weight of 0.1. For the MoE model, we retain
the pretrained routing configuration and apply a global-batch auxiliary
load-balancing loss with a coefficient of $10^{-3}$. We use AdamW with a
weight decay of 0.1. The learning rate is linearly warmed up for the first
3\% of training steps and then cosine-decayed to zero. For offline-packed
32K-token sequences, we enforce sample-level isolation using per-sample
cumulative sequence boundaries. These boundaries induce block-diagonal
causal attention, reset positional indices and Gated DeltaNet states, and
mask next-token targets that cross sample boundaries. More details are in Table~\ref{tab:qwen35_midtraining}.

\begin{table*}[t]
\centering
\small
\renewcommand{\arraystretch}{1.08}
\setlength{\tabcolsep}{7pt}

\begin{tabularx}{\textwidth}{
    @{}
    >{\raggedright\arraybackslash}X
    >{\raggedright\arraybackslash}p{0.29\textwidth}
    >{\raggedright\arraybackslash}p{0.29\textwidth}
    @{}
}
\toprule
\textbf{Hyperparameter} &
\textbf{Qwen3.5-9B} &
\textbf{Qwen3.5-35B-A3B} \\
\midrule

\rowcolor{black!6}[0pt][0pt]
\multicolumn{3}{@{}c@{}}{\textbf{Model}} \\

Initialization
& Qwen3.5-9B-Base
& Qwen3.5-35B-A3B-Base \\

Finetuning type
& full parameter
& full parameter \\

Training backend
& Megatron-Bridge
& Megatron-Bridge \\

Precision
& bf16 mixed
& bf16 mixed \\

Attention backend
& auto, cuDNN fused preferred
& auto, cuDNN fused preferred \\

Vision modules
& frozen
& frozen \\

MTP layers
& 1
& 1 \\

MTP loss coefficient
& 0.1
& 0.1 \\

\addlinespace[3pt]

\rowcolor{black!6}[0pt][0pt]
\multicolumn{3}{@{}c@{}}{\textbf{Data}} \\

Training objective
& causal LM on all valid tokens
& causal LM on all valid tokens \\

Valid training tokens
& 33,699,997,413
& 33,699,997,413 \\

Physical training tokens
& 34,133,245,952
& 34,133,245,952 \\

Fixed packed rows
& 1,041,664
& 1,041,664 \\

Sequence length
& 32,768
& 32,768 \\

Packing efficiency
& 98.73\%
& 98.73\% \\

Pre-packed THD sequences
& enabled
& enabled \\

Shuffle
& global random cyclic sampler
& global random cyclic sampler \\

\addlinespace[3pt]

\rowcolor{black!6}[0pt][0pt]
\multicolumn{3}{@{}c@{}}{\textbf{Optimization}} \\

Epochs
& 1 (8,138 steps)
& 1 (8,138 steps) \\

Micro batch size
& 1 packed row
& 1 packed row \\

Gradient accumulation
& 16
& 32 \\

Global batch size
& 128 packed rows
& 128 packed rows \\

Physical token batch
& 4,194,304 tokens/step
& 4,194,304 tokens/step \\

Valid token batch
& $\sim$4.14M tokens/step
& $\sim$4.14M tokens/step \\

Optimizer
& distributed Adam
& distributed Adam \\

Peak learning rate
& $5 \times 10^{-5}$
& $5 \times 10^{-5}$ \\

Minimum learning rate
& 0
& 0 \\

LR scheduler
& cosine
& cosine \\

Warmup
& 244 steps (3\%)
& 244 steps (3\%) \\

Weight decay
& 0.1
& 0.1 \\

Adam betas
& $(0.9, 0.95)$
& $(0.9, 0.95)$ \\

Adam epsilon
& $1 \times 10^{-8}$
& $1 \times 10^{-8}$ \\

Gradient clipping
& 1.0
& 1.0 \\

Per-token loss
& enabled
& enabled \\

\addlinespace[3pt]

\rowcolor{black!6}[0pt][0pt]
\multicolumn{3}{@{}c@{}}{\textbf{Parallelism}} \\

Number of GPUs
& $4 \times 8$ H200
& $4 \times 8$ H200 \\

TP / PP / CP / DP
& 4 / 1 / 1 / 8
& 4 / 2 / 1 / 4 \\

EP / ETP
& 1 / 4
& 4 / 1 \\

Sequence parallelism
& enabled
& enabled \\

Parameter gather overlap
& enabled
& enabled \\

Gradient reduce overlap
& enabled
& enabled \\

\addlinespace[3pt]

\rowcolor{black!6}[0pt][0pt]
\multicolumn{3}{@{}c@{}}{\textbf{MoE}} \\

Number of experts
& N/A
& 256 \\

Experts per token
& N/A
& 8 \\

Token dispatcher
& N/A
& all-to-all \\

Grouped GEMM
& N/A
& enabled \\

Router precision
& N/A
& fp32 \\

Load balancing
& N/A
& global auxiliary loss \\

Auxiliary loss coefficient
& N/A
& $1 \times 10^{-3}$ \\

\addlinespace[3pt]

\rowcolor{black!6}[0pt][0pt]
\multicolumn{3}{@{}c@{}}{\textbf{Kernels}} \\

Bias-activation fusion
& enabled
& enabled \\

Masked-softmax fusion
& enabled
& enabled \\

Cross-entropy fusion
& enabled
& enabled \\

MoE permutation fusion
& N/A
& enabled \\

\addlinespace[3pt]

\bottomrule
\end{tabularx}

\caption{Qwen3.5 mid-training configurations on 32 NVIDIA H200 GPUs.}
\label{tab:qwen35_midtraining}
\end{table*}

\subsection{SFT Training Details}
The SFT training details are provided in Table~\ref{tab:qwen35_native_mcore_sft}. The Qwen3 model is trained using the same configuration as Qwen3.5-9B.

\begin{table*}[t]
\centering
\small
\renewcommand{\arraystretch}{1.08}
\setlength{\tabcolsep}{8pt}

\begin{tabularx}{\textwidth}{
    @{}
    >{\raggedright\arraybackslash}X
    >{\raggedright\arraybackslash}p{0.28\textwidth}
    >{\raggedright\arraybackslash}p{0.28\textwidth}
    @{}
}
\toprule
\textbf{Hyperparameter} &
\textbf{Qwen3.5-9B} &
\textbf{Qwen3.5-35B-A3B} \\
\midrule

\rowcolor{black!6}[0pt][0pt]
\multicolumn{3}{@{}c@{}}{\textit{\textbf{Model}}} \\

Initialization
& Qwen3.5-9B-Base
& Qwen3.5-35B-A3B-Base \\

Training backend
& Slime + Native MCore
& Slime + Native MCore \\

Precision
& bf16
& bf16 \\

Attention implementation
& MCore flash backend
& MCore flash backend \\

Vision modules
& frozen
& frozen \\

MTP layers
& 1
& 1 \\

MTP loss coefficient
& 0.1
& 0.1 \\

\addlinespace[3pt]

\rowcolor{black!6}[0pt][0pt]
\multicolumn{3}{@{}c@{}}{\textit{\textbf{Data}}} \\

Training samples
& 35,861
& 35,861 \\

Maximum sequence length
& 147,456 (144K)
& 147,456 (144K) \\

Maximum position embeddings
& 262,144
& 262,144 \\

Mean sequence length
& 36,445 tokens
& 36,445 tokens \\

P95 sequence length
& 84,446 tokens
& 84,446 tokens \\

Shuffle
& global shuffle
& global shuffle \\

\addlinespace[3pt]

\rowcolor{black!6}[0pt][0pt]
\multicolumn{3}{@{}c@{}}{\textit{\textbf{Optimization}}} \\ 

Epochs
& 5
& 5 \\

Per-device batch size
& dynamic
& dynamic \\

Maximum tokens per GPU
& 131,072
& 131,072 \\

Gradient accumulation
& dynamic to reach GBS
& dynamic to reach GBS \\

Global batch size
& 128 samples
& 128 samples \\

Effective token batch
& $\sim$4.665M tokens/step
& $\sim$4.665M tokens/step \\

Optimizer
& distributed Adam
& distributed Adam \\

Peak learning rate
& $3 \times 10^{-5}$
& $4 \times 10^{-5}$ \\

Minimum learning rate
& $4 \times 10^{-6}$
& $4 \times 10^{-6}$ \\

LR scheduler
& cosine
& cosine \\

Warmup
& 50 steps ($\sim$3.6\%)
& 50 steps ($\sim$3.6\%) \\

Weight decay
& 0.01
& 0.01 \\

Adam betas
& $(0.9, 0.98)$
& $(0.9, 0.98)$ \\

Gradient clipping
& 1.0
& 1.0 \\

Attention dropout
& 0.0
& 0.0 \\

Hidden dropout
& 0.0
& 0.0 \\

Cross-entropy
& linear fused CE
& linear fused CE \\

\addlinespace[3pt]

\rowcolor{black!6}[0pt][0pt]
\multicolumn{3}{@{}c@{}}{\textit{\textbf{Parallelism}}} \\

Number of GPUs
& $4 \times 8$ H200
& $4 \times 8$ H200 \\

TP / PP / CP / DP
& 4 / 1 / 1 / 8
& 4 / 1 / 1 / 8 \\

EP / ETP
& 1 / 1
& 8 / 1 \\

Sequence parallelism
& enabled
& enabled \\

MoE dispatcher
& N/A
& flex + DeepEP \\

MoE routing
& N/A
& softmax, top-8 of 256 \\

MoE load balancing
& N/A
& global auxiliary loss \\

MoE auxiliary coefficient
& N/A
& $1 \times 10^{-3}$ \\

\addlinespace[3pt]

\bottomrule
\end{tabularx}

\caption{Native MCore full-parameter SFT configurations for Qwen3.5 models.}
\label{tab:qwen35_native_mcore_sft}
\end{table*}

\clearpage
\section{Resource Usage of Memory Compaction}
\label{sec:compaction_resource_usage}

\textbf{Generation usage.}
We report generation and verification resource usage across 26,708 recorded turns. Table~\ref{tab:compaction_resource_usage} summarizes the compaction events and cumulative input and output tokens for these records. Input tokens are summed over each call's actual context, accounting for history replacement after compaction and including retained checkpoints in subsequent calls. Output tokens include reasoning, response text, and tool calls.

\begin{table}[htbp]
\centering
\caption{Recorded resource usage for memory-compaction evaluation. Input and output token counts are in millions.}
\label{tab:compaction_resource_usage}
\begin{tabular}{lrrr}
\toprule
Dataset & Compaction events & Input tokens & Output tokens \\
\midrule
MiniF2F & 104 & 414.29 & 22.44 \\
FATE-M & 33 & 140.67 & 7.89 \\
FATE-H & 365 & 1,646.51 & 63.48 \\
\midrule
\textbf{Total} & \textbf{502} & \textbf{2,201.47} & \textbf{93.81} \\
\bottomrule
\end{tabular}
\end{table}

\textbf{Compaction overhead and accounting checks.}
All 502 compactions use rule-based extraction without additional LLM calls for summarization. Token counts for all 1,004 pre- and post-compaction context snapshots match the historical server-side records. Thus, the input accounting includes the context actually consumed after each history replacement.

\textbf{Lean verification.}
Recorded verification-call durations sum to approximately 11.02 hours, with a median of 0.734 seconds per call; 91.4\% of calls take at most 10 seconds. These cumulative durations include overlapping concurrent calls and should not be interpreted as end-to-end elapsed time or CPU-hours.

\textbf{Scope.}
Memory compaction is an auxiliary analysis; our main contribution is agent-native mid-training and its benefits for downstream agentic proving. 

\end{document}